\documentclass{article}

\usepackage{PRIMEarxiv}
\usepackage[utf8]{inputenc}
\usepackage[T1]{fontenc}
\usepackage{hyperref}
\usepackage{url}
\usepackage{booktabs}
\usepackage{amsfonts}
\usepackage{nicefrac}
\usepackage{microtype}
\usepackage{lipsum}
\usepackage{fancyhdr}
\usepackage{graphicx}
\graphicspath{{media/}}
\usepackage{multirow}
\usepackage{algorithmic}
\usepackage{algorithm}
\usepackage{makecell}
\usepackage{amsmath}
\usepackage{microtype}
\usepackage{graphicx}
\usepackage{subfigure}
\usepackage{booktabs}
\usepackage{marvosym}
\usepackage{natbib}
\usepackage{utfsym}

\usepackage{hyperref}
\hypersetup{hidelinks, colorlinks=true} 

\title{Taylor-Sensus Network: Embracing Noise \\ to Enlighten Uncertainty for Scientific Data}

\author{
  Guangxuan Song, Dongmei Fu \\
  University of Science and Technology Beijing \\
  Beijing\\
  \texttt{fdm\_ustb@ustb.edu.cn} \\
  \And
  Zhongwei Qiu \\
  Zhejiang University \\
  Hangzhou\\
  \AND
  Jintao Meng \\
  National Key Laboratory of Security Communication \\
  Chengdu \\
  \And
  Dawei Zhang \\
  University of Science and Technology Beijing \\
  Beijing \\
}

\begin{document}
\maketitle

\begin{abstract}
Uncertainty estimation is crucial in scientific data for machine learning. 
Current uncertainty estimation methods mainly focus on the model's inherent uncertainty, while neglecting the explicit modeling of noise in the data.
Furthermore, noise estimation methods typically rely on temporal or spatial dependencies, which can pose a significant challenge in structured scientific data where such dependencies among samples are often absent.
To address these challenges in scientific research, we propose the Taylor-Sensus Network (TSNet). TSNet innovatively uses a Taylor series expansion to model complex, heteroscedastic noise and proposes a deep Taylor block for aware noise distribution. 
TSNet includes a noise-aware contrastive learning module and a data density perception module for aleatoric and epistemic uncertainty.
Additionally, an uncertainty combination operator is used to integrate these uncertainties, and the network is trained using a novel heteroscedastic mean square error loss.
TSNet demonstrates superior performance over mainstream and state-of-the-art methods in experiments, highlighting its potential in scientific research and noise resistance. It will be open-source to facilitate the community of ``AI for Science''.
\end{abstract}


\section{Introduction}
\label{introduction}

Exploring scientific data has made rapid progress with the rise of ``AI for Science''\citep{wang2023scientific}, especially in the field of machine learning (ML). 
However, the uncertainty in scientific data poses both risks and opportunities for ML. ML based on erroneous data, without thoroughly exploring data uncertainty, can lead to biased or incorrect conclusions \citep{liu2016rethinking}. On the other hand, data uncertainty provides insights for evaluating hypotheses and related experiments, accelerating the exploration of scientific domains \citep{coley2019robotic}. Therefore, modeling uncertainty in scientific data is crucial in the ML process. 
However, methods for uncertainty face many challenges, especially when dealing with complex scientific data.

\begin{figure}[t]
    \begin{center}
    \centerline{\includegraphics[width=0.6\columnwidth]{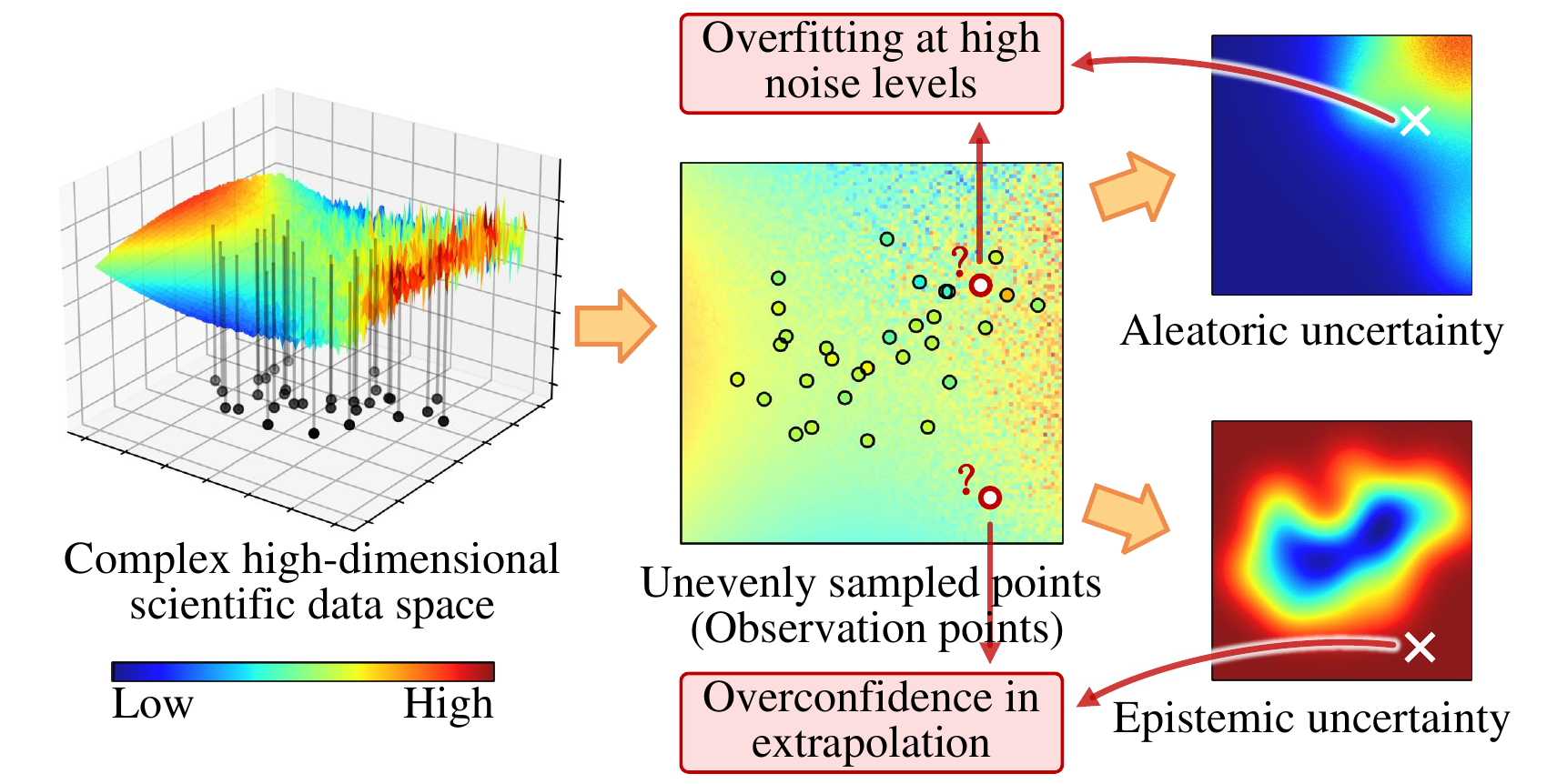}}
    \caption{In scientific ML, aleatoric and epistemic uncertainties can lead to issues such as overfitting and overconfidence. However, they also provide valuable insights for data exploration.}
    \label{fig:intro}
    \end{center}
    \vspace{-0.5cm}
\end{figure}

Figure \ref{fig:intro} shows that data uncertainty arises from two main sources: aleatoric and epistemic uncertainty \citep{hullermeier2021aleatoric, sluijterman2024evaluate}.
Aleatoric uncertainty, attributed to intrinsic data noise \citep{emadi2023decision}, is often considered inevitable and, if unaddressed, can result in overfitting, especially in spaces with high noise.
Despite the potential for statistical analysis to quantify this noise through repeated experiments, practical constraints - chiefly the prohibitive cost - limit this approach, thereby highlighting the challenge of estimating noise in scientific data due to informational scarcity. Epistemic uncertainty arises from data sparsity and non-uniform sampling. In many scientific experiments, it is unrealistic to obtain a large number of dense data points due to objective limitations \citep{gao2022innovative, stephany2024pde}. This can hinder the model's comprehensive understanding of the scientific data space, leading to overconfidence in predictions beyond the training data distribution.
The presence of these uncertainties not only affects the accuracy of data analysis but can also lead to overlooking valuable exploration spaces in scientific research \citep{ghahramani2015probabilistic}. 
However, the advantage of deep learning in capturing data features has led to many developments in the study of data uncertainty, providing opportunities to fully exploit existing scientific data and estimate its uncertainty.

To address uncertainty estimation, dominant methods include Bayesian techniques and ensemble methods \citep{abdar2021review, hoffmann2021uncertainty}, which largely focus on model uncertainties or variations but do not explicitly incorporate mechanisms for data noise.
Denoising research provides strategies for explicit noise estimation, mainly in temporal series data (e.g. audio, sensor time series) \citep{zheng2022denoising} and spatial data (e.g. images) \citep{chang2020spatial}, as well as their combination in videos\citep{song2022tempformer}, exploiting temporal and spatial dependencies of features for noise reduction. 
However, they do not account for uncertainty, and the scientific data on which this paper focuses does not have clear temporal or spatial dependencies, presenting a significant challenge in assessing noise levels.

Given the limitations of current methods in accounting for data noise as a source of uncertainty and quantifying uncertainty in scientific data, we propose the Taylor-Sensus Network (TSNet) for noise and information density estimation.
Our method addresses both aleatoric and epistemic uncertainties, which are related to noise distribution parameters and sampling variability, respectively. 
To address aleatoric uncertainty, we use a heteroscedastic noise transformation based on Taylor series expansion theory and propose a Deep Taylor Block (DTB) for noise awareness.
Additionally, we provide a theoretical explanation of how data augmentation enhances model robustness from an uncertainty perspective, as a particular case of the DTB.
Furthermore, we propose the Noise-Aware Contrastive Learning Module (NCL) to improve noise pattern recognition and the Data Density Perception Module (DPM) for estimating epistemic uncertainty through density perception. 
We then combine these uncertainties using the Uncertainty Combination Operator (UCO) and employ the reparameterization trick alongside a heteroscedastic MSE loss to address both types of uncertainties, enabling robust estimation under limited information.
TSNet utilizes the foundation of Taylor expansion theory and explicit awareness of noise and information density to provide a more reliable measure of uncertainty.

The effectiveness of TSNet was validated on toy examples and public scientific datasets. 
The method's noise resilience and uncertainty estimation capabilities were validated through anti-noise and active learning experiments.

TSNet's theoretical innovation enhances the precision and reliability of data analysis, thereby deepening scientific insights. Our contributions can be summarized as follows:
\begin{itemize}
\setlength{\parskip}{0pt}
    \item We present a novel heteroscedastic noise transformation based on the Taylor expansion theory. This simplifies the impact of noise and provides an explanation for data augmentation with added noise, laying the theoretical foundation for TSNet.
    \item We propose a novel framework called TSNet, for estimating uncertainty. It includes a DTB designed based on heteroscedastic noise transformation and provides a more accurate awareness of uncertainty through contrastive learning and density perception.
    \item We propose a novel heteroscedasticity MSE loss for learning multivariate Gaussian distributions. This enhances the model's ability to process complex data.
    \item  Extensive and diverse experiments demonstrate the efficacy of TSNet, elucidating its working principle and highlighting its potential for scientific data analysis. TSNet outperforms state-of-the-art methods in experiments, underscoring its superior capability in managing uncertainties and noise.
\end{itemize}

TSNet will be open-sourced to facilitate the ``AI for Science'' community and promote scientific advancement.

\section{Related Work}

We study the uncertainty estimation for scientific data and try to quantify uncertainty by estimating noise. Here, we briefly review the related works on uncertainty estimation and denoising methods.

\subsection{Uncertainty Estimation}
Uncertainty estimation often uses Bayesian techniques, such as Markov chain Monte Carlo (MCMC) \citep{lampinen2001bayesian, blundell2015weight, papamarkou2022challenges} and variational inference (VI) \citep{gunapati2022variational, lu2024forecasting}, as well as ensemble methods \citep{abdar2021review}.
MCMC can theoretically estimate posterior distributions accurately but is time-consuming. VI is a more efficient method than MCMC but relies on solving optimization problems and provides only approximations of posterior distributions with limited flexibility.
Monte Carlo (MC) dropout \citep{zheng2021uncertainty, choubineh2023applying} is a practical approximation within modern deep learning frameworks. 
However, these methods mainly capture model uncertainty and have limitations in handling complex data noise patterns. 
Ensemble methods \citep{yang2020reliability}, which infer uncertainty from model discrepancies, show robustness to some data variability but lack explicit mechanisms for identifying noise.
The computational complexities of MC approximations and the methods' inability to directly address data noise and its distribution limit their utility in scientific data analysis.

\subsection{Denoising Methods}
Uncertainty estimation and denoising are related fields. 
In the visual domain, \citep{zhang2023mm, qiu2023learning} learn noise distributions from noisy-clean sample pairs. 
In scientific contexts, however, ``clean'' data acquisition is challenging, but \citep{lehtinen2018noise2noise} gets around this by using noise-only pairs, suggesting that noise features can reveal underlying noise patterns. 
In addition, denoising efforts on data with temporal and spatial dependencies (e.g. images, and time series data) provide reference points for noise identification through feature dependencies \citep{rasul2021autoregressive, baptista20221d}. 
However, these methods may not be suitable for structured scientific data lacking such dependencies, nor can they measure noise uncertainty.

\section{Preliminaries}
This section defines the noise estimation problem by considering both feature noise and system noise in scientific data lacking feature dependencies, in order to estimate uncertainty. We derive the heteroscedastic noise transformation process to decouple noise, which also explains the effectiveness of noise perturbation in data augmentation and provides a theoretical foundation for our method.

\subsection{Problem Definition}
To facilitate the formulation of the equations, we first define the noise in scientific datasets. A scientific dataset typically consists of pairs of samples $(\textbf{X},\textbf{Y}_n)$, where $\textbf{X}$ is the ideal state of the data features and $\textbf{Y}_n$ is the label with noise. In the real-world scientific process, the difference between the real state of the research subject, $\textbf{X}_n$, and the recorded $\textbf{X}$ is considered as feature noise, $N_i$, represented as $\textbf{X}_n = \textbf{X} + N_i$. In addition, the recorded observation $\textbf{Y}_N$ deviates from the actual state $\textbf{Y}$ due to the system noise $N_o$, resulting in a gap.

We consider the additive noise model in scientific data:
\begin{equation}
\label{eq:noiseMod}
\left\{
\begin{array}{l}
\begin{aligned}
\textbf{Y}_n = & F(\textbf{X}_n)+N_o  
    = Noise(F(\textbf{X})) \\
\textbf{X}_n = & \textbf{X} + N_i
\end{aligned}
\end{array},
\right.
\end{equation}
where $\textbf{X}_n, \textbf{X} \in \mathbb{R}^m$ and $\textbf{Y}_n, \textbf{Y} \in \mathbb{R}^l$. $m$ and $l$ represent the dimensions of the features and observations respectively, $N_i$ and $N_o$ correspond to feature noise and system noise, respectively.
$F( \cdot ) $ represents the objective operational laws of an ideal scientific system, while $Noise(\cdot)$ represents the interference caused by subjective human factors, technical limitations, and practical conditions, etc. 
The noise conforms to the sampling $N_i \sim \mathcal{D}_{N_i}$, $N_o \sim \mathcal{D}_{N_o}$. 
Observation points $\textbf{X} \sim \mathcal{D}_{sample}$ indicate that the expected observation points follow a certain distribution $\mathcal{D}_{sample}$. 

We define aleatoric uncertainty $U_a \propto {g(N_i, N_o)}$, where $g(\cdot)$ describes the process from the distribution parameters of $\mathcal{D}_{N_i}$ and $\mathcal{D}_{N_o}$ to $U_a$, with $U_a \in \mathbb{R}^l$.
Similarly, epistemic uncertainty $U_e \propto {h(\textbf{X})}$, where $h(\cdot)$ describes the mapping from the parameters of $\mathcal{D}_{sample}$ to $U_e$, with $U_e \in \mathbb{R}^l$.
Additionally, the comprehensive uncertainty $U_c = U_a + U_e$.

We define uncertainty estimating as learning a mapping
\begin{equation}
    \Psi: \textbf{X} \to (\textbf{Y}, U_c),
\end{equation}
which predicts the ideal observations and uncertainty based on feature data.

\begin{figure}[t!]
    \begin{center}
    \centerline{\includegraphics[width=0.85\linewidth]{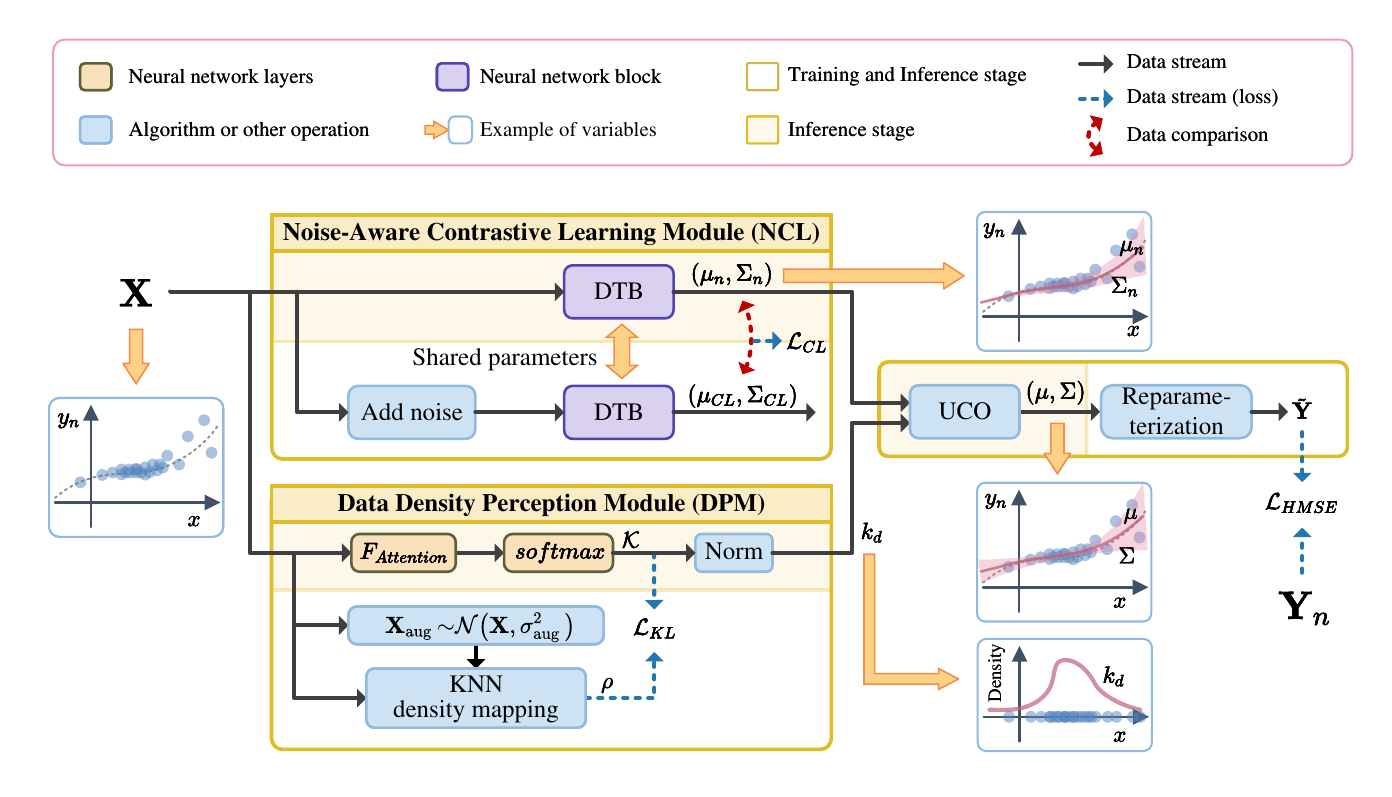}}
    \caption{The Taylor-Sensus Network (TSNet) framework. It integrates Noise-Aware Contrastive Learning Module (NCL), Data Density Perception Module (DPM), and Uncertainty Combination Operator (UCO). The Deep Taylor Block (DTB) in NCL is crucial for estimating the $(\mu_{n}, \Sigma_{n})$ parameters. And NCL introduces re-noising for feature $\textbf{X}$ to derive $(\mu_{CL}, \Sigma_{CL})$ via the shared-parameters DTB, which improves noise learning via $\mathcal{L}_{CL}$. DPM uses KL loss $\mathcal{L}_{KL}$ for density estimation $\mathcal{K}$ against KNN density mapping $\rho$, yielding density-sensitive weight $k_d$. UCO refines $(\mu_{n}, \Sigma_{n})$ to $(\mu, \Sigma)$ using $k_d$. The reparameterization trick generates $\widetilde{Y}$ under $\mathcal{N}(\mu, \Sigma)$ for heteroscedastic MSE loss $\mathcal{L}_{HMSE}$. $X_{aug} \sim \mathcal{N}(X, \sigma^2_{aug})$ represents data augmentation by sampling from the distribution $\mathcal{N}(X, \sigma^2_{aug})$.}
    \label{fig:framework}
    \end{center}
\end{figure}

\subsection{Heteroscedastic Noise Transformation}
Real-world noise characteristics are often complex. To address this, we use a heteroscedastic noise model that depends on $\textbf{X}$, which poses unique challenges in uncertainty estimating. Additionally, we assume a heteroscedastic Gaussian distribution to derive the noise distribution parameters, expressed as $\mathcal{N} (0, \Sigma^2(x))$.

In the presence of heteroscedastic noise, the noise model defined by Equation (\ref{eq:noiseMod}) is $\textbf{Y}_n = F(\textbf{X} + N_i(\textbf{X})) + N_o(\textbf{X})$, which introduces input-output uncertainty due to the combined influence of $F(\cdot)$ and noise, creating significant non-linearity. 
Directly fitting this noise model with neural networks is challenging. To elucidate the effect of noise on the function, we perform a heteroscedastic noise transformation. We start by expanding $F(\textbf{X} + N_i(\textbf{X}))$ at $\textbf{X}$ using a multivariate Taylor series, resulting in
\begin{equation}
\label{eq:Taylor}
\begin{aligned}
F(\textbf{X}+N_i(\textbf{X})) = & F(\textbf{X}) + ( \nabla F(\textbf{X}) ) ^\top N_i(\textbf{X}) + \\
         & \frac{1}{2!} N_i(\textbf{X}) ^\top \textbf{H}_F(\textbf{X}) N_i(\textbf{X}) + o^n
\end{aligned},
\end{equation}
where $\nabla F(\textbf{X})$ and $\textbf{H}_F(\textbf{X})$ are the gradient and Hessian matrix of $F(\cdot)$ at $\textbf{X}$, respectively. The calculation is described in the \ref{Apx:cmteqTaylor}. When the noise is small, keeping only the first order term gives
\begin{equation}
\label{eq:Taylor_res}
\textbf{Y}_n(\textbf{X}) \simeq F(\textbf{X}) + ( \nabla F(\textbf{X}) ) ^\top N_i(\textbf{X}) + N_o(\textbf{X}).
\end{equation}
Equation (\ref{eq:Taylor_res}) quantifies the impact of noise on the function. Letting $N_{ft}(\textbf{X}) = ( \nabla F(\textbf{X}) ) ^\top N_i(\textbf{X})$, $N_{ft}(\textbf{X})$ depicts the influence of feature noise on the observation at $\textbf{X}$.

When the noise follows a heteroscedastic Gaussian distribution, we have
\begin{equation}
    N_i(\textbf{X}) \sim \mathcal{N}(0, \Sigma_i(\textbf{X})); \quad N_o(\textbf{X}) \sim \mathcal{N}(0, \Sigma_o(\textbf{X})).
\end{equation}
Assuming independence between $N_i(\textbf{X})$ and $N_o(\textbf{X})$, and utilizing the linearity property of the multivariate Gaussian distribution, it follows that
\begin{equation}
\label{eq:gasdis}
    N(\textbf{X}) \sim \mathcal{N}(0,  (\nabla F) ^\top \Sigma_i(\textbf{X}) \nabla F + \Sigma_o(\textbf{X})),
\end{equation}
where $N(\textbf{X})=N_{ft}(\textbf{X}) + N_o(\textbf{X})$. For a detailed derivation, see \ref{apx:cmptEqMultigas}.

Heteroscedastic noise transformation simplifies functions with inherent noise into linear forms, facilitating neural network construction by allowing the network to learn the primary behavior of the function and to be aware of noise levels. Additionally, it quantifies the effect of feature noise on function output, which is essential for awareness of aleatory uncertainty in data-scarce scenarios.

\subsection{Data Uncertainty Insights into Data Augmentation}

The addition of noise perturbations to neural network inputs serves as a common data augmentation strategy, 
widely used in fields such as computer vision \citep{shorten2019survey, xu2023comprehensive} and data processing \citep{lashgari2020data, jiang2023noisemol, garcea2023data}. 
We try to explain the efficacy of data augmentation through the Taylor series expansion. 
Considering a function $y=f(x)$, data augmentation is the introduction of a noise term $\delta$ to $x$. 
By expanding $f(x+\delta)$ around $x$ using a multivariate Taylor series, as in Equation (\ref{eq:Taylor}), we derive
\begin{equation}
f(x+\delta) = f(x) + ( \nabla f(x) ) ^\top \delta + 
          \frac{1}{2!} \delta ^\top \textbf{H}_f(x) \delta + o^n .
\end{equation}
Define $f_{\delta}(x) = ( \nabla f(x) ) ^\top \delta + \frac{1}{2!} \delta ^\top \textbf{H}_f(x) \delta + o^n$, which captures the effect of data augmentation noise $\delta$. During training, the neural network learns a mixture of $f(x)$ and $f_{\delta}(x)$. Ideally, the network detects and learns the critical aspects of $f(x)$ while mitigating the influence of $\delta$ (i.e. $f_{\delta}(x)$). This strategy enhances the network's ability to understand the intrinsic structure of the input data and to remain robust to noise.

From the perspective of heteroscedastic noise transformation, data augmentation is a form of uncertainty estimation. Unlike this type of data augmentation, which aims to reduce the influence of noise, our approach - based on the inherent noise of scientific datasets and Taylor series theory - aims to teach neural networks to learn data relationships while quantifying noise levels, thereby enriching scientific data exploration with more informative results.

\section{Methods}

To handle the uncertainty estimation problems for scientific data, we propose a framework that estimates aleatoric uncertainty through noise-aware contrastive learning and explicitly models epistemic uncertainty through data density perception. Based on heteroscedastic noise transformation, we propose the Deep Taylor Block, which, supported by the theory of multivariate Taylor series, aids in more accurate data noise estimation. We have designed a learning mechanism for the framework that enables efficient learning without compromising theoretical precision.

\subsection{Framework}

The framework of the Taylor-Sensus Network (TSNet) is shown in Figure \ref{fig:framework}. We propose a Deep Taylor Block (DTB) for predicting outputs while estimating heteroscedastic noise distribution. The Noise-Aware Contrastive Learning Module (NCL) of TSNet employs the shared-parameters DTB to conduct contrastive learning, capturing data patterns and aleatoric uncertainty. 
The Data Density Perception Module (DPM) learns the epistemic uncertainty of training samples. 
An Uncertainty Combination Operator (UCO) integrates the two types of uncertainties to yield the final estimation. 
Using the reparameterization trick, TSNet ensures network trainability and achieves end-to-end learning through its DTB, NCL, DPM, and UCO components.

\begin{figure}[t!]
    \begin{center}
    \centerline{\includegraphics[width=0.6\linewidth]{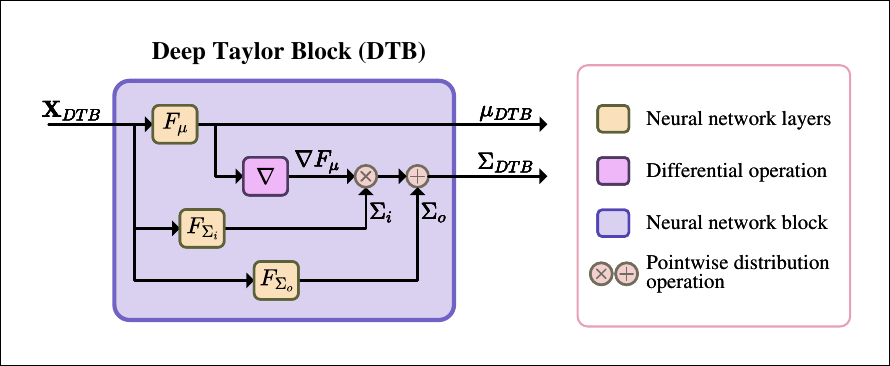}}
    \caption{The Deep Taylor Block (DTB). According to heteroscedastic noise transformation (Equation (\ref{eq:Taylor_res}) and (\ref{eq:gasdis})), $F_{\mu}(\cdot)$, $F_{\Sigma_i}(\cdot)$, and $F_{\Sigma_o}(\cdot)$ predict the noise-free data label $\mu_{DTB}$, the feature noise distribution parameters $\Sigma_i(\textbf{X}_{DTB})$, and the system noise distribution parameters $\Sigma_o(\textbf{X}_{DTB})$, respectively.}
    \label{fig:DTB}
    \end{center}
\end{figure}

\begin{figure}[t!]
    \begin{center}
    \centerline{\includegraphics[width=0.75\linewidth]{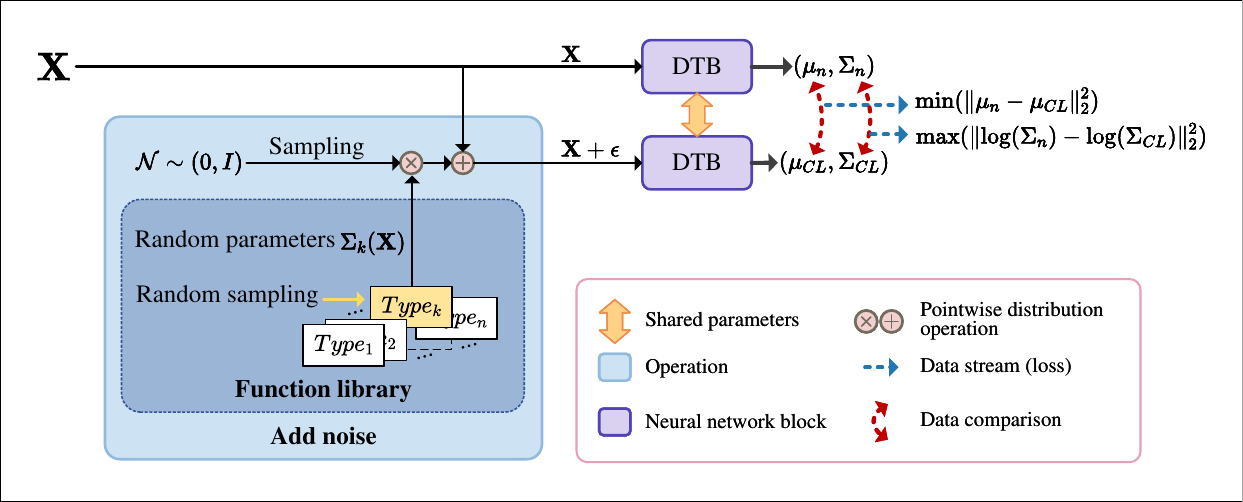}}
    \caption{The Noise-Aware Contrastive Learning Module (NCL).
    Zero-mean noise $\epsilon$ following the distribution $\mathcal{N}(0, \Sigma_k(\mathbf{X}))$ is randomly introduced to the sample features $\mathbf{X}$. The noisy features $\mathbf{X} + \epsilon$ and the original sample $\mathbf{X}$ are both passed through the DTB to predict their respective means and variances. By minimizing the mean and maximizing the variance in this process, contrastive learning is achieved, thereby enhancing the perception of feature noise.}
    \label{fig:NCL}
    \end{center}
\end{figure}

\subsection{Deep Taylor Block}
The Deep Taylor Block (DTB) is proposed for data modeling and awareness of aleatoric uncertainty. The DTB is designed based on heteroscedastic noise transformation as
\begin{equation}
    \label{eq:DTB}
    \left\{
    \begin{array}{l}
    \begin{aligned}
        \mu_{DTB} = & F_{\mu}(\textbf{X}) \\
        \Sigma_{DTB} = &\left( \nabla F_{\mu}(\textbf{X}) \right)^\top F_{\Sigma_i}(\textbf{X}) \left( \nabla F_{\mu}(\textbf{X}) \right) + F_{\Sigma_o}(\textbf{X})
    \end{aligned}
    \end{array}.
    \right.
\end{equation}
As shown in Figure \ref{fig:DTB}, the DTB consists of three subnetworks
\begin{equation}
    \begin{array}{l}
    \begin{aligned}
        F_{\mu}(\textbf{X}_{DTB}; \theta_{\mu}) = \mu_{DTB}\\
        F_{\Sigma_i}(\textbf{X}_{DTB}; \theta_{\Sigma_i}) = \Sigma_i\\
        F_{\Sigma_o}(\textbf{X}_{DTB}; \theta_{\Sigma_o}) = \Sigma_o
    \end{aligned}
    \end{array},
\end{equation}
where $\theta_{\mu}$, $\theta_{\Sigma_i}$, and $\theta_{\Sigma_o}$ are the learnable network parameters. $F_{\mu}(\cdot)$, $F_{\Sigma_i}(\cdot)$, and $F_{\Sigma_o}(\cdot)$ predict the noise-free data label $\mu_{DTB}$, the feature noise distribution parameters $\Sigma_i(\textbf{X}_{DTB})$, and the system noise distribution parameters $\Sigma_o(\textbf{X}_{DTB})$, respectively. 
The $\nabla$ denotes the first-order differentiation of $F_{\mu}(\cdot)$. In modern deep learning frameworks, automatic differentiation (e.g., Autograd in PyTorch and GradientTape in TensorFlow) typically eliminates the need for manually deriving the $\nabla F_{\mu}(\cdot)$ expression and leverages the frameworks' optimized computations.
The errors from the higher-order terms omitted in Equation (\ref{eq:Taylor_res}) will also be estimated in $F_{\Sigma_o}(\cdot)$.

Due to the presence of \(\nabla F_{\mu}(\cdot)\) during forward propagation, second-order derivatives of \(F_{\mu}(\cdot)\) must be computed during the backward pass of the loss function. To avoid learning failure, \(F_{\mu}(\cdot)\) must be twice differentiable. To ensure generality, in this paper, \(F_{\mu}(\cdot)\), \(F_{\Sigma_i}(\cdot)\), and \(F_{\Sigma_o}(\cdot)\) all employ multilayer perceptrons (MLPs) with sigmoid activation functions.
The DTB decouples data modeling and noise distribution parameters estimation, using specialized subnetworks to improve accuracy and enhance the interpretability of the model.

\subsection{Noise-Aware Contrastive Learning Module}

Inspired by the effectiveness of contrastive learning in data representation, we propose the Noise-Aware Contrastive Learning Module (NCL) to enhance the noise awareness of neural networks. Contrastive learning has proven to be a powerful technique for learning robust feature representations by contrasting positive and negative pairs of samples. Leveraging this concept, the NCL module is designed to make neural networks more sensitive to noise, thereby improving their robustness and generalization capabilities.

As depicted in Figure \ref{fig:NCL}, NCL consists of two parameter-sharing DTBs that process the original features $\mathbf{X}$ and the same features with added heteroscedastic noise centered on the zero mean. Specifically, zero-mean noise $\epsilon$ following the distribution $\mathcal{N}(0, \Sigma_k(\mathbf{X}))$ is randomly introduced to the sample features $\mathbf{X}$ using the reparameterization trick \citep{kingma2014auto}. The construction of the random parameter $\Sigma_k(\mathbf{X})$ involves sampling a mathematical function (e.g., constant, trigonometric, exponential, etc.) from a function library and generating its parameters randomly, thus ensuring the diversity of $\Sigma_k$.

This process yields noise distribution parameters for $\mathcal{N}(\mu_n, \Sigma_n)$ and $\mathcal{N}(\mu_{CL}, \Sigma_{CL})$, respectively. Since zero-mean heteroscedastic noise does not alter the mean of the data but affects the covariance of its uncertainty, the loss function of NCL is designed to reflect this characteristic:
\begin{equation}
    \mathcal{L}_{CL} = \lambda_{CL1} \|\mu_n - \mu_{CL}\|^2_2 - \lambda_{CL2} \|\text{log}(\Sigma_n) - \text{log}(\Sigma_{CL})\|^2_2,
\end{equation}
where $\lambda_{CL1}$ and $\lambda_{CL2}$ are hyperparameters, and $\|\cdot\|_2$ denotes the Euclidean distance. By leveraging contrastive learning, TSNet can enhance the capacity and generalization of the DTB through improved noise awareness.

\subsection{Data Density Perception Module}

We propose the Data Density Perception Module (DPM), which improves TSNet's understanding of the data sample distribution by using a neural network to learn from the augmented data features. 
To improve density perception in sparsely sampled data spaces, DPM employs perturbative sampling along each feature dimension, creating augmented features $X_{aug} \sim \mathcal{N}(\textbf{X}, \sigma^2_{aug})$, where $\sigma^2_{aug}$ is determined by the variance of the data feature distribution. 
To computer the density of a data point $x \in \textbf{X} \cup \textbf{X}_{aug}$ in feature space, K-Nearest Neighbors (KNN) density mapping is used, which uses KNN and normalizes the density to a probability distribution $\rho(\textbf{X})$ with the softmax function:
\begin{equation}
    \rho(\textbf{X}) = softmax\left(
    \sum \nolimits_{i \in \mathcal{I}(x, K)} (\|x - x_i\|_2^2)^{-1}
    \right),
\end{equation}
where 
$\mathcal{I}(x, K) = \left\{ \text{argmin}_{1 \leq i \leq \tau} 
\begin{matrix}
\{\|x - x_i\|_2^2\}_{(1)}^{(K)}
\end{matrix} \right\}$, where $x_i \in \textbf{X}$, $\tau$ is the total number of data points.

To perceive the density $\mathcal{K}$ of test sample points in the original data feature space $\textbf{X}$ during inference, the neural network approximates the KNN density mapping as
\begin{equation}
    softmax(F_{Attention}(\textbf{X}; \theta_{\mathcal{K}})) = \mathcal{K},
\end{equation}
where $\theta_{\mathcal{K}}$ are learnable parameters.
The loss function is computed using the Kullback-Leibler divergence:
\begin{equation}
    \mathcal{L}_{KL}(\mathcal{K} \| \rho) = \sum \mathcal{K}(\textbf{X}) \log \frac{\mathcal{K}(\textbf{X})}{\rho(\textbf{X})}.
\end{equation}
DPM can internalize the density distribution of data through data augmentation and KNN density mapping without the need for explicit density labels. This approach theoretically enables the model to better understand the global data distribution, facilitating the estimation of epistemic uncertainty.

\subsection{Comprehensive Uncertainty and Learning}

The Uncertainty Combination Operator (UCO) combines the normalized density $\mathcal{K}$ from DPM, represented as $k_d$, with the prior influence $\mathcal{N}(\mu_p, \Sigma_p)$ into the NCL predicted distribution $\mathcal{N}(\mu_n, \Sigma_n)$ to produce the comprehensive uncertainty $\mathcal{N}(\mu, \Sigma)$. Assuming independence between $\mathcal{N}(\mu_n, \Sigma_n)$ and $\mathcal{N}(\mu_p, \Sigma_p)$, and considering heteroscedasticity, the following expressions are derived:
\begin{equation}
\label{eq:finallres}
    \left\{
    \begin{array}{l}
    \begin{aligned}
        \mu(\textbf{X}) = & k_d(\textbf{X}) \mu_n(\textbf{X}) + (1 - k_d(\textbf{X})) \mu_p(\textbf{X}) \\
        \Sigma(\textbf{X}) = & k_d^2(\textbf{X}) \Sigma_n(\textbf{X}) + (1 - k_d(\textbf{X}))^2 \Sigma_p(\textbf{X}) 
    \end{aligned}
    \end{array},
    \right.
\end{equation}
as detailed in \ref{apx:cmptfinallres}.
Equation (\ref{eq:finallres}) describes the predicted data label and comprehensive uncertainty, respectively, as outputs of TSNet. 

During the learning stage, we use the reparameterization trick \citep{kingma2014auto} to sample from $\mathcal{N}(\mu,\Sigma)$:
\begin{equation}
    \widetilde{\textbf{Y}}=\mu(\textbf{X})+\Sigma \otimes \epsilon,
\end{equation}
where $\epsilon \sim \mathcal{N}(0, \textbf{I})$ is random noise from a standard Gaussian distribution, $\otimes$ denotes element-wise multiplication, and $\textbf{I}$ is the identity matrix. 

Considering the heteroscedasticity of the noise, we propose a heteroscedastic MSE loss,
\begin{equation}
\label{eq:HMSE}
    \mathcal{L}_{HMSE} = \lambda_{H} \sum \log|\Sigma| + \sum (y_n - \widetilde{y})^T \Sigma^{-1} (y_n - \widetilde{y}),
\end{equation}
as derived in \ref{apx:cmptHMSE}, where $|\cdot|$ denotes the determinant calculation, and $\Sigma^{-1}$ is the inverse matrix of $\Sigma$, $\lambda_{H}$ is hyperparameters. 

Computing the determinants and inverses of matrices is typically computationally intensive. However, considering the heteroscedasticity, using a diagonal covariance matrix directly captures the essential information effectively. The multivariate Gaussian distribution dictates that $\Sigma(X)$ is positive definite and symmetric. At point $X$, we perform an eigenvalue decomposition of $\Sigma(X)$:
\begin{equation}
    \Sigma(X) = V(X)\Lambda(X)V(X)^{\top},
\end{equation}
where $\Lambda(X)$ is the diagonal matrix of eigenvalues, and $V(X)$ is the corresponding matrix of orthogonal eigenvectors. By applying the local linear transformation $V(X)^{\top}$, we can transform $\Sigma(X)$ into a coordinate system based on its principal components, making each dimension independent. This transformation enables TSNet to assume that $\Sigma(X)$ is diagonal while still effectively representing the complex dependencies of data, thereby avoiding the computational complexity of computing determinants and inverses.

The total loss function of TSNet is described as
\begin{equation}
    \mathcal{L} = \lambda_{CL} \mathcal{L}_{CL} + \lambda_{KL} \mathcal{L}_{KL} + \mathcal{L}_{HMSE},
\end{equation}
where, $\lambda_{CL}$ and $\lambda_{KL}$ are coefficients of the components.

\begin{algorithm}[t!]
    \caption{TSNet Algorithm}
    \label{alg:TSNet}
    \textbf{Input}
    $(\textbf{X},\textbf{Y}_n)$: dataset. 
    
    \textbf{Parameter}
    $\mu_p$, $\Sigma_p$: prior distribution parameters of epistemic uncertainty.
    $T$: maximum iteration.
        
    \textbf{Output}
    $\mu$,$\Sigma$: parameters of predicted distribution.
    
    \textbf{Training stage:}
    \begin{algorithmic}[1]
        \FOR {epoch = 1 to T}
            \STATE $(\mu_{n}, \Sigma_{n}) \leftarrow DTB(\textbf{X})$
            \STATE $(\mu_{CL}, \Sigma_{CL}) \leftarrow DTB(addNoise(\textbf{X}))$
            \STATE $\mathcal{L}_{CL} = \|\mu_n - \mu_{CL}\|^2_2 - \|\text{log}(\Sigma_n) - \text{log}(\Sigma_{CL})\|^2_2$
            \STATE $\rho(\textbf{X}), \mathcal{K} \leftarrow DPM(\textbf{X})$
            \STATE $\mathcal{L}_{KL}(\mathcal{K} \| \rho) = \sum \mathcal{K}(\textbf{X}) \log \frac{\mathcal{K}(\textbf{X})}{\rho(\textbf{X})}$
            \STATE $k_d = norm(\mathcal{K})$
            \STATE $\mu \leftarrow k_d \mu_n + (1 - k_d) \mu_p$; $\Sigma \leftarrow k_d^2 \Sigma_n + (1 - k_d)^2 \Sigma_p$
            \STATE $\widetilde{\textbf{Y}} \leftarrow \mu(\textbf{X})+\Sigma \odot \epsilon$
            \STATE $\mathcal{L}_{HMSE} = \lambda_{H} \log|\Sigma| + (y - \widetilde{y})^T \Sigma^{-1} (y - \widetilde{y})$
            \STATE Update the model w.r.t. $\mathcal{L} = \mathcal{L}_{CL} + \mathcal{L}_{KL} + \mathcal{L}_{HMSE}$
        \ENDFOR
        \STATE \textbf{return} {$\mu$, $\Sigma$}
    \end{algorithmic}
    
    \textbf{Inference stage:}
    \begin{algorithmic}[1]
    \footnotesize{
        \STATE Same as lines 2, 5, 7, 8, and 13 in the training stage.
    }
    \end{algorithmic}
\end{algorithm}

\subsection{Algorithm}
The training and inference stages are shown in Algorithm \ref{alg:TSNet}. In the training stage, the input feature $\textbf{X}$ is processed through the DTB, NCL, DPM, and UCO to obtain an uncertainty distribution. 
Reparameterization generates $\widetilde{\textbf{Y}}$, and loss $\mathcal{L}$ is calculated for parameter optimization. 
In the inference stage, $\textbf{X}$ is processed through DTB and DPM to obtain aleatoric and epistemic uncertainties, which are then combined by UCO. The TSNet, which avoids Monte Carlo sampling by directly estimating the noise distribution parameters, speeds up both training and inference.

\begin{table}[t!]
    \caption{Experimental results for scientific datasets. The results are reported as 'mean ± standard deviation'. The best results are highlighted in bold.}
    \label{tab:exp2}
    \begin{center}
    \resizebox{0.85\linewidth}{!}{

    \renewcommand{\arraystretch}{1.2}
    
    \begin{tabular}{c | r@{±}l r@{±}l r@{±}l | r@{±}l r@{±}l r@{±}l | r@{±}l r@{±}l r@{±}l}
    \hline
    
    \hline
    Datasets & \multicolumn{6}{c|}{Matbench steels} & \multicolumn{6}{c|}{Diabetes} & \multicolumn{6}{c}{Forest fires} \\
    \hline
    Sample count & \multicolumn{6}{c|}{312} & \multicolumn{6}{c|}{442} & \multicolumn{6}{c}{517} \\
    \hline
    \multirow{2}{*}{Methods} & \multicolumn{6}{c|}{Testing set} & \multicolumn{6}{c|}{Testing set} & \multicolumn{6}{c}{Testing set} \\
    \cline{2-19}
     & \multicolumn{2}{c}{MSE(×10$^4$)$\downarrow$} & \multicolumn{2}{c}{MAE(×10)$\downarrow$} & \multicolumn{2}{c|}{NLL$\downarrow$} & \multicolumn{2}{c}{MSE(×10$^3$)$\downarrow$} & \multicolumn{2}{c}{MAE(×10)$\downarrow$} & \multicolumn{2}{c|}{NLL$\downarrow$} & \multicolumn{2}{c}{MSE(×10$^2$)$\downarrow$} & \multicolumn{2}{c}{MAE(×10)$\downarrow$} & \multicolumn{2}{c}{NLL$\downarrow$} \\
    \hline
    GPRMK & 3.63 & 0.82 & 12.21 & 0.63 & 5.98 & 0.12 & 4.89 & 0.54 & 5.60 & 0.39 & 5.25 & 0.02 & 1.79 & 0.17 & 1.18 & 0.06 & 3.79 & 0.06 \\
    MCRF & 1.97 & 0.27 & 10.11 & 0.61 & 7.33 & 0.07 & 3.44 & 0.34 & 4.75 & 0.24 & 6.87 & 0.03 & 1.68 & 0.11 & 1.16 & 0.04 & 6.40 & 0.02 \\
    MLP & 3.93 & 1.67 & 11.26 & 1.66 & \multicolumn{2}{c|}{N/A} & 3.56 & 0.29 & 4.75 & 0.15 & \multicolumn{2}{c|}{N/A} & 1.79 & 0.32 & 1.24 & 0.05 & \multicolumn{2}{c @{}}{N/A} \\
    MC-drop & 4.64 & 1.03 & 15.66 & 1.30 & 6.91 & 0.26 & 3.71 & 0.26 & 5.00 & 0.26 & 6.70 & 0.38 & 1.81 & 0.22 & 1.19 & 0.08 & 6.58 & 0.56 \\
    BNN-ELBO & 1.97 & 0.73 & 9.97 & 0.84 & 6.98 & 0.68 & 3.59 & 0.67 & 4.73 & 0.35 & 6.88 & 0.13 & 1.96 & 0.48 & 1.13 & 0.13 & 39.44 & 0.16 \\
    BNN-MCMC & 3.05 & 0.17 & 12.98 & 0.60 & 9.41 & 0.03 & 4.07 & 0.51 & 5.10 & 3.73 & 5.13 & 0.01 & 1.64 & 0.22 & 1.20 & 0.01 & 3.72 & 0.00\\
    BBP & 3.58 & 1.48 & 11.26 & 1.53 & 6.70 & 0.05 & 5.12 & 0.41 & 4.70 & 0.23 & 5.92 & 0.82 & 1.71 & 0.13 & 1.22 & 0.06 & 3.90 & 0.03 \\
    VAER & 4.85 & 0.43 & 14.59 & 0.82 & \multicolumn{2}{c|}{N/A} & 7.06 & 0.24 & 5.44 & 0.25 & \multicolumn{2}{c|}{N/A} & 1.73 & 0.18 & 1.22 & 0.05 & \multicolumn{2}{c @{}}{N/A} \\
    Evidential & 2.00 & 0.36 & 9.97 & 0.96 & 6.90 & 0.57 & 4.00 & 0.27 & 5.13 & 1.22 & 11.24 & 0.27 & 1.65 & 0.18 & 1.14 & 0.11 & 10.78 & 0.15 \\
    MT-ENet & 1.88 & 0.19 & 9.35 & 4.45 & 6.18 & 0.34 & 3.81 & 0.77 & 4.90 & 0.42 & 9.49 & 0.50 & 1.65 & 0.35 & 1.16 & 0.90 & 5.04 & 0.26 \\
    \hline
    \textbf{TSNet(Ours)} & \textbf{1.45} & \textbf{0.12} & \textbf{8.54} & \textbf{0.42} & \textbf{5.67} & \textbf{0.32} & \textbf{3.20} & \textbf{0.26} & \textbf{4.49} & \textbf{0.16} & \textbf{5.06} & \textbf{0.33} & \textbf{1.62} & \textbf{0.09} & \textbf{1.12} & \textbf{0.03} & \textbf{3.61} & \textbf{0.06}\\
    \hline
    
    \hline
    \end{tabular}
    }
    
    \vspace{0.2cm}
    
    \resizebox{0.85\linewidth}{!}{

    \renewcommand{\arraystretch}{1.2}
    
    \begin{tabular}{ c | r@{±}l r@{±}l r@{±}l | r@{±}l r@{±}l r@{±}l | r@{±}l r@{±}l r@{±}l }
    \hline
    
    \hline
    Datasets & \multicolumn{6}{c|}{Concrete compressive strength} & \multicolumn{6}{c|}{Wine quality-red} & \multicolumn{6}{c}{Wine quality-white} \\
    \hline
    Sample count & \multicolumn{6}{c|}{1030} & \multicolumn{6}{c|}{1599} & \multicolumn{6}{c}{4898} \\
    \hline
    \multirow{2}{*}{Methods} & \multicolumn{6}{c|}{Testing set} & \multicolumn{6}{c|}{Testing set} & \multicolumn{6}{c}{Testing set} \\
    \cline{2-19}
     & \multicolumn{2}{c}{MSE(×10)$\downarrow$} & \multicolumn{2}{c}{MAE$\downarrow$} & \multicolumn{2}{c|}{NLL$\downarrow$} & \multicolumn{2}{c}{MSE(×0.01)$\downarrow$} & \multicolumn{2}{c}{MAE(×0.1)$\downarrow$} & \multicolumn{2}{c|}{NLL$\downarrow$} & \multicolumn{2}{c}{MSE(×0.1)$\downarrow$} & \multicolumn{2}{c}{MAE$\downarrow$} & \multicolumn{2}{c}{NLL$\downarrow$} \\
    \hline
    GPRMK & 4.26 & 0.79 & 4.57 & 0.27 & 4.26 & 0.14 & 6.99 & 0.28 & 1.65 & 0.06 & -0.25 & 0.03 & 1.97 & 0.07 & 2.98 & 0.04 & 0.02 & 0.01 \\
    MCRF & 2.55 & 0.56 & 3.52 & 0.38 & 3.02 & 0.02 & 6.10 & 0.51 & 1.60 & 0.03 & -0.54 & 0.07 & 0.93 & 0.04 & 2.07 & 0.05 & -0.12 & 0.10 \\
    MLP & 4.03 & 1.19 & 3.97 & 0.27 & \multicolumn{2}{c|}{N/A} & 7.11 & 0.62 & 1.73 & 0.03 & \multicolumn{2}{c|}{N/A} & 0.78 & 0.03 & 1.94 & 0.02 & \multicolumn{2}{c}{N/A} \\
    MC-drop & 6.18 & 0.53 & 6.31 & 0.19 & 3.16 & 0.09 & 6.86 & 0.54 & 1.79 & 0.05 & 1.46 & 0.31 & 1.31 & 0.09 & 3.10 & 0.11 & 0.13 & 0.01 \\
    BNN-ELBO & 2.56 & 0.42 & 3.65 & 0.28 & 3.36 & 0.27 & 6.66 & 0.29 & 1.72 & 0.02 & 0.86 & 0.46 & 0.93 & 0.09 & 2.00 & 0.05 & -0.22 & 0.01 \\
    BNN-MCMC & 3.94 & 0.16 & 3.90 & 0.37 & 3.56 & 0.02 & 6.76 & 0.28 & 1.74 & 0.04 & -0.13 & 0.03 & 1.93 & 0.05 & 2.75 & 0.16 & 0.57 & 0.00 \\
    BBP & 3.42 & 0.47 & 3.51 & 0.10 & 3.84 & 0.00 & 6.98 & 0.62 & 1.70 & 0.04 & -0.05 & 0.06 & 1.29 & 0.07 & 2.88 & 0.07 & 1.04 & 0.02 \\
    VAER & 4.10 & 0.75 & 4.14 & 0.41 & \multicolumn{2}{c|}{N/A} & 7.38 & 0.11 & 1.83 & 0.03 & \multicolumn{2}{c|}{N/A} & 1.09 & 0.09 & 2.09 & 0.11 & \multicolumn{2}{c}{N/A} \\
    Evidential & 2.50 & 0.26 & 3.37 & 0.22 & 3.33 & 0.07 & 7.87 & 0.34 & 1.78 & 0.04 & -0.21 & 0.79 & 0.92 & 0.04 & 1.94 & 0.30 & -0.58 & 0.05 \\
    MT-ENet & 2.49 & 0.40 & 3.44 & 0.22 & 3.30 & 0.04 & 7.34 & 0.30 & 1.81 & 0.03 & -0.32 & 0.10 & 0.96 & 0.03 & 1.99 & 0.14 & -0.31 & 0.06 \\
    \hline
    \textbf{TSNet(Ours)} & \textbf{2.47} & \textbf{0.37} & \textbf{3.20} & \textbf{0.27} & \textbf{3.00} & \textbf{0.08} & \textbf{5.28} & \textbf{0.25} & \textbf{1.56} & \textbf{0.04} & \textbf{-0.59} & \textbf{0.06} & \textbf{0.72} & \textbf{0.05} & \textbf{1.81} & \textbf{0.02} & \textbf{-0.62} & \textbf{0.08}\\
    \hline
    
    \hline
    \end{tabular}
    
    }
    \end{center}
    \vspace{-0.5cm}
\end{table}

\section{Experiment}

\subsection{Experimental Setup}

Our objective was to assess the efficacy of TSNet in scientific data, marked by inherent noise and sparsity. We conducted comprehensive experiments on both real-world scientific datasets and two toy examples. The experiments were conducted on an NVIDIA GTX 1080Ti.

\subsubsection{Scientific Datasets}
The scientific datasets exemplify typical challenges in scientific data analysis, such as noise and non-uniform sampling. 
Six UCI public datasets from the fields of materials science, medicine, environmental science, and food science were used. These include Diabetes \citep{efron2004least}, Matbench Steels \citep{dunn2020benchmarking}, Concrete Compressive Strength \citep{misc_concrete_compressive_strength_165}, Forest Fires \citep{misc_forest_fires_162}, and Wine (red/white) Quality \citep{misc_wine_quality_186}. 
These datasets demonstrate the models' applicability across various scientific contexts. 
For details on data preprocessing, please refer to \ref{apx:data_pre}.

Despite the limited data sizes typically found in real-world scientific settings, we tested the performance of TSNet in the efficiency and scalability evaluation using the large Year-Prediction-MSD  \citep{misc_year_prediction_msd_203} dataset (514K) from the UCI public dataset collection.

\begin{figure}[t!]
    \begin{center}
    \centerline{\includegraphics[width=\linewidth]{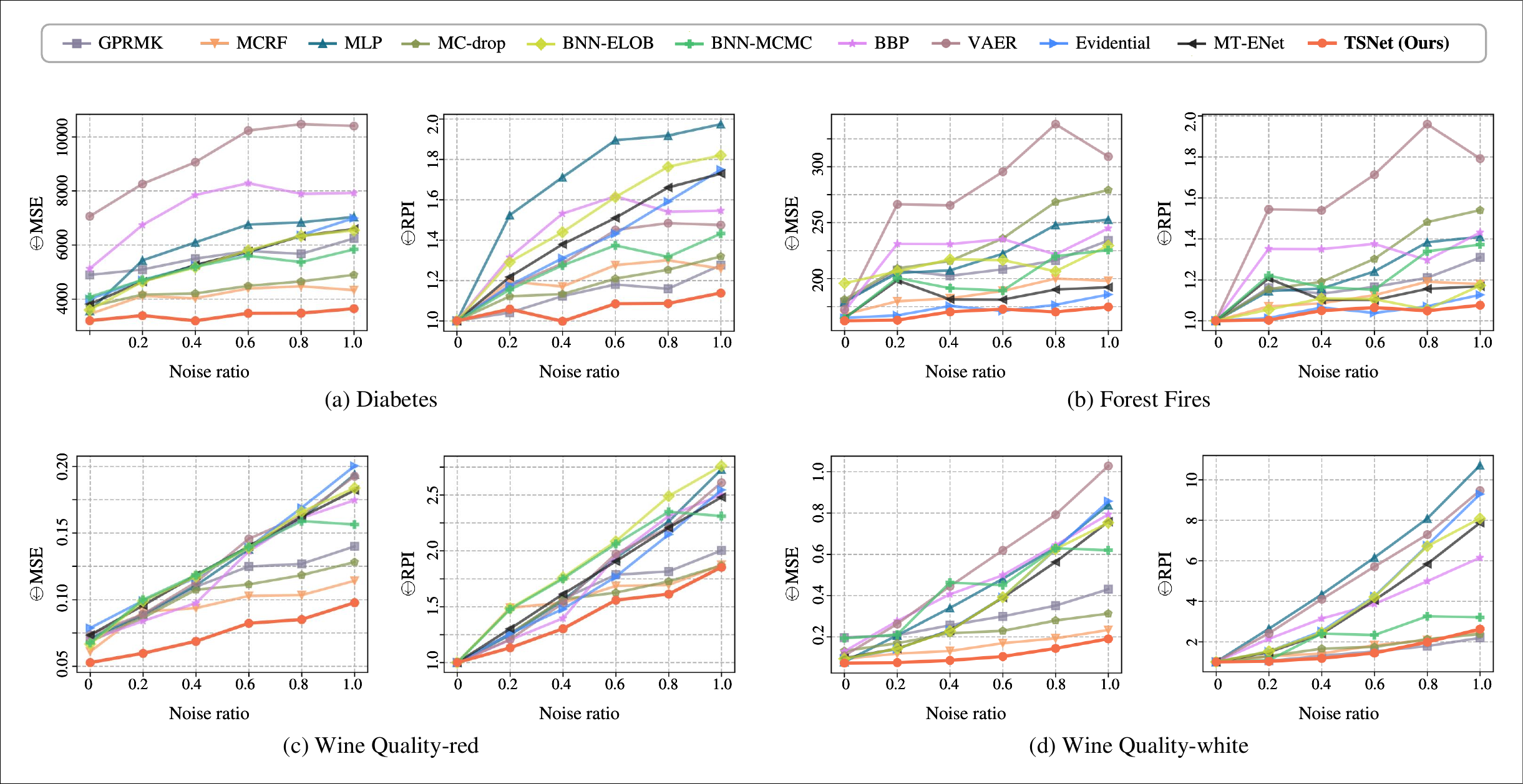}}
    \caption{Experimental results of the anti-noise experiments. Relative Performance Index (RPI) is calculated as $\text{MSE}_i / \text{MSE}_0$, quantifying models' normalized performance change under different noise ratios. (a)\textasciitilde(g) shows the results on the different datasets.\textcircled{$\downarrow$}indicates that lower evaluation results are preferable.}
    \label{fig:exp4}
    \end{center}
    \vspace{-0.5cm}
\end{figure}

\subsubsection{Benchmark}
TSNet was benchmarked against several mainstream and state-of-the-art regression and uncertainty estimation methods, including Gaussian process regression with multiple kernels (GPRMK) \citep{chugh2019trading}, Markov Chain Random Forest (MCRF) \citep{yang2020reliability}, multilayer perceptron (MLP) \citep{zhang2023applications},  MC dropout (MC-drop) \citep{folgoc2021mc, choubineh2023applying}, BNN-ELBO \citep{wei2023multi}, BNN-MCMC \citep{zhang2020cyclical}, Bayes by Backprop (BBP) \citep{blundell2015weight}, VAE-based regression model (VAER) \citep{zhao2019variational, cai2020detecting}, Evidenceial \citep{amini2020deep} and MT-ENet \citep{oh2022improving}.

\subsubsection{Evaluation Metrics} 

The performance of the models was evaluated using well-established metrics: Mean Squared Error (MSE), Mean Absolute Error (MAE), and Negative Log-Likelihood (NLL). These metrics provide comprehensive insights into different aspects of model performance. MSE measures the average squared difference between the predicted and actual values, highlighting larger errors due to the squaring function. MAE calculates the average absolute difference, providing a more interpretable metric in terms of the magnitude of errors. NLL, on the other hand, evaluates the model's uncertainty estimation by considering both the predicted mean and variance. The formulas for these metrics are as follows:
\begin{equation}
    \begin{array}{l}
    \begin{aligned}
        \text{MSE} = \frac{1}{n} \sum_{i=1}^n (y_i - \hat{y}_i)^2 \\
        \text{MAE} = \frac{1}{n} \sum_{i=1}^n |y_i - \hat{y}_i| \\
        \text{NLL} = \log(\sigma) + \frac{(y-\mu)^2}{2\sigma^2}
    \end{aligned}
    \end{array}.
\end{equation}
In the NLL formula, the inconsistent rate of change between \( \log(\sigma) \) and \( \frac{1}{\sigma^2} \) may affect the validity of the assessment. 
Despite this issue, NLL is still used as a reference metric in this study due to its prevalence in uncertainty estimation tasks for uncertainty estimation.

\subsection{Experiments on Scientific Datasets}

To demonstrate the generality and applicability of TSNet across various scientific domains, we conducted a series of experiments using publicly available scientific datasets. These datasets were selected to represent a range of real-world scenarios where uncertainty estimation is crucial.

\subsubsection{Comparison Experiment}
In the scientific dataset experiments, each dataset was divided into training, validation, and test sets in a 6:2:2 ratio. This division was performed three times with random shuffling to ensure robust results. All models underwent training and testing on these splits, maintaining comparability. While we cannot access noise-free scientific data to test models, our use of diverse datasets, appropriate evaluation metrics, and rigorous experimental setups allows for a relative performance comparison among models. 

Table \ref{tab:exp2} shows TSNet's superiority in managing uncertainties and noise in scientific data, achieving optimal results in MSE, MAE, and NLL. This indicates its high generalizability. 
Notably, TSNet achieves state-of-the-art (SOTA) performance across these metrics, highlighting its efficacy in handling complex noise and providing reliable uncertainty estimates.
TSNet's prediction of uncertainty enhances data comprehension and improves accuracy on test sets.
Furthermore, the lack of ideal noise-free conditions in scientific research makes our scientific dataset experiment, which approximates real-world applications, a valuable demonstration of TSNet's effectiveness in addressing these challenges, showing its practical value in scientific applications.

\begin{figure}[t!]
\begin{center}
\centerline{\includegraphics[width=\linewidth]{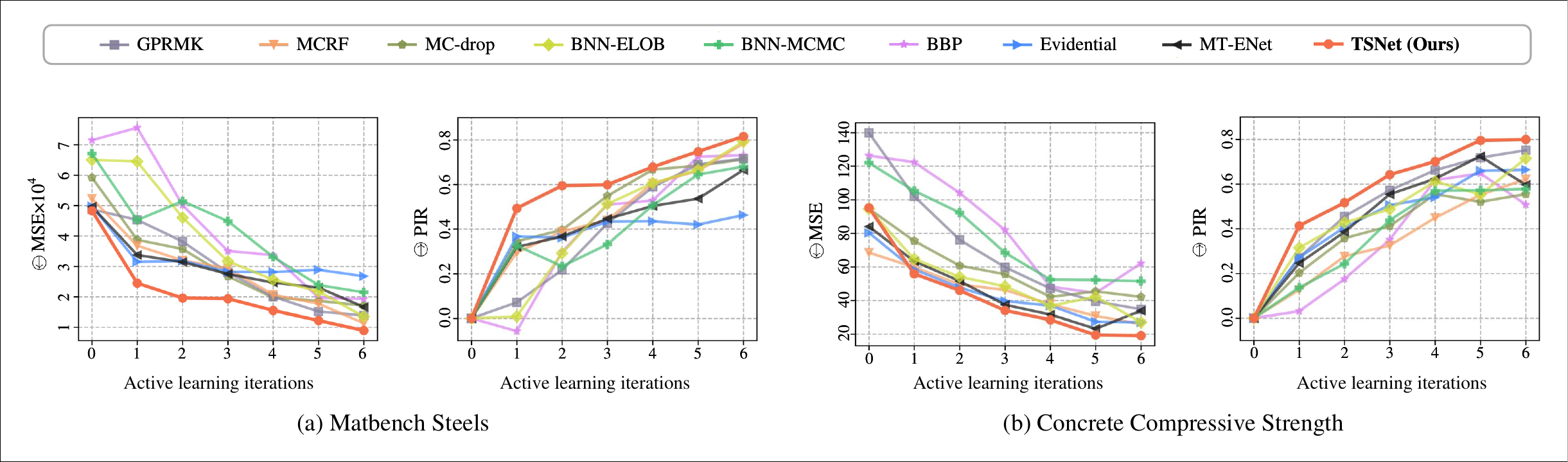}}
\caption{Experimental results of the active learning experiments. The Performance Improvement Ratio (PIR) calculated as $(MSE_0 - MSE_i) / MSE_0$, measures the improvement over active learning iterations. (a) and (b) show the results on the different datasets. \textcircled{$\downarrow$} and \textcircled{$\uparrow$}, indicate that, respectively, lower or higher evaluation results are preferable.}
\label{fig:exp5}
\vspace{-0.5cm}
\end{center}
\end{figure}

\subsubsection{Anti-noise Experiments}

To evaluate the robustness of TSNet in noisy data, we introduced noise to varying rates of the training data entries across four datasets and evaluated the models' performance on the testing set. Noise was added to 0\%, 20\%, 40\%, 80\%, and 100\% of the data entries after normalization. The feature noise followed a normal distribution $\mathcal{N}\sim(0, 0.4)$ and the system noise followed $\mathcal{N}\sim(0, 0.8)$, while the testing set remained unchanged to assess the model's noise resistance.

TSNet consistently achieved the lowest MSE across all evaluated noise levels, as shown in Figure \ref{fig:exp4}. Due to TSNet's initially lower MSE, fluctuations in MSE had a greater impact on the Relative Performance Index (RPI), defined as $\text{RPI} = {{MSE}_i}/{{MSE}_0}$, which occasionally resulted in TSNet not achieving the top position in terms of RPI. Nevertheless, TSNet demonstrated superior noise resistance, achieving the lowest RPI in 15 out of 20 noisy experiments and close-to-lowest in the remaining ones.

This experiment underscores TSNet's adaptability to high-noise environments, validating its efficacy and reliability in real-world conditions characterized by noisy data. It highlights TSNet's ability to handle noisy scientific data, indicating its significant potential in this area.

\subsubsection{Active Learning Experiments}
Active learning has been applied in domains where data acquisition is expensive or limited, such as materials science \citep{schmidt2019recent, nugraha2020mesoporous, liu2022experimental}. To test the effectiveness of TSNet in an active learning setting, we conducted experiments on the Matbench Steels and Concrete Compressive Strength datasets. These datasets were chosen due to their relevance in materials science and the high cost associated with obtaining large amounts of labeled data.

The experiments on active learning followed a structured process, as described by \citep{yang2016active}. Initially, all models were trained on the same initial dataset. In each cycle, the models were trained on this data and the uncertainty of the remaining samples was evaluated. The top 10\% of samples with the highest uncertainty were then added to the training set for the next cycle. This iterative process was repeated to simulate scientific progress with limited data, allowing us to assess the impact of uncertainty-based sample selection on model performance.
To ensure a comprehensive evaluation, the experiment tested multiple models including GPRMK, MCRF, MC-drop, BNN-ELBO, BNN-MCMC, BBP, Evidential, MT-ENet, and TSNet, using uncertainty estimates as the basis for sample selection.

Figure \ref{fig:exp5} shows that TSNet demonstrated the most significant reduction in MSE compared to other methods. Except for the initial iteration on the Concrete Compressive Strength dataset, where TSNet did not achieve the lowest MSE, it consistently yielded the lowest predictive MSE throughout the subsequent cycles. This consistent performance underscores TSNet's proficiency in selecting informative samples via uncertainty estimation.
Additionally, TSNet achieved the highest Performance Improvement Ratio (PIR), defined as $(\text{MSE}_0 - \text{MSE}_i) / \text{MSE}_0$, across all experiments. This indicates TSNet's effectiveness in identifying high-value data points, thereby rapidly enhancing the understanding of the sample space. These results suggest that TSNet not only accelerates the learning process but also significantly improves model efficacy, making it highly advantageous in active learning scenarios with limited data.

\subsection{Experiment of Efficiency and Scalability Evaluation}

\begin{table}[t]
    \caption{Resource utilization on large-scale dataset (Year-Prediction-MSD).}
    \label{tab:exp_RU}
    \begin{center}
    \resizebox{0.55\linewidth}{!}{
        \renewcommand{\arraystretch}{1.2}
        \begin{tabular}{c|c|c}
        \hline

        \hline
        \multicolumn{2}{c|}{\textbf{Metric}} & \textbf{TSNet(Ours)} \\
        \hline
        \multicolumn{2}{c|}{Parameters} & 3.018 M \\
        \hline
        \multirow{2}{*}{Training} & Total time (Training set size = 463715) & 101.03 ± 2.59 s \\
        & Max GPU memory usage & 0.995 ± 0.000 GB \\
        \hline
        \multirow{2}{*}{Inference} & Total time (Testing set size = 51630) & 0.411 ± 0.036 s \\
        & Max GPU memory usage & 0.341 ± 0.001 GB \\
        \hline

        \hline
        \end{tabular}

    }

    \end{center}
\end{table}

\begin{figure}[t!]
\begin{center}
\centerline{\includegraphics[width=0.6\linewidth]{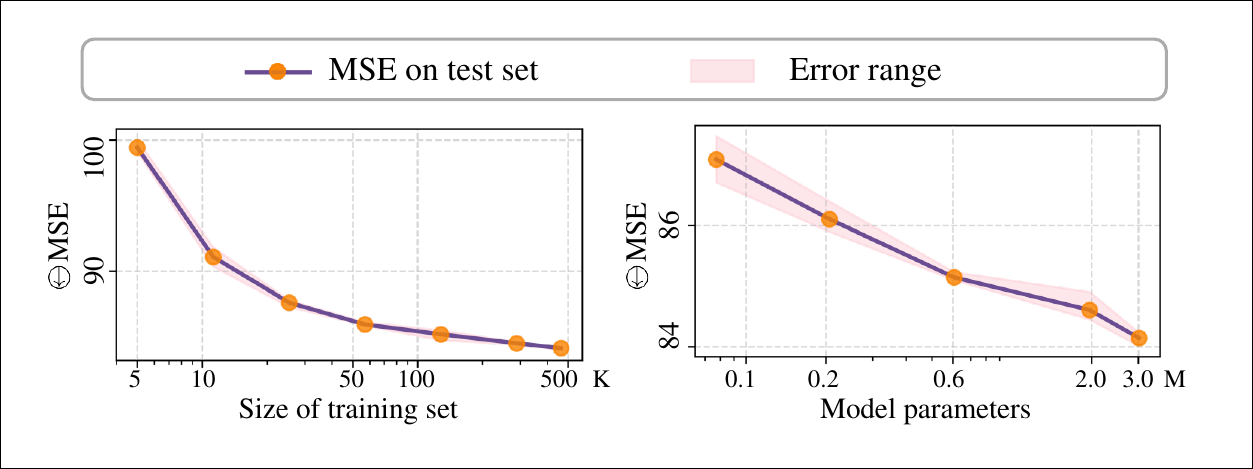}}
\caption{
Experimental results of TSNet scalability. 
The TSNet framework does not hinder model scalability, on the contrary, it shows the potential for increased accuracy with large-scale datasets and substantial parameter sizes. \textcircled{$\downarrow$}indicates that lower evaluation results are preferable.}
\label{fig:exp5-6}
\end{center}
\vspace{-0.7cm}
\end{figure}

\begin{figure}[t]
    \begin{center}
    \centerline{\includegraphics[width=\linewidth]{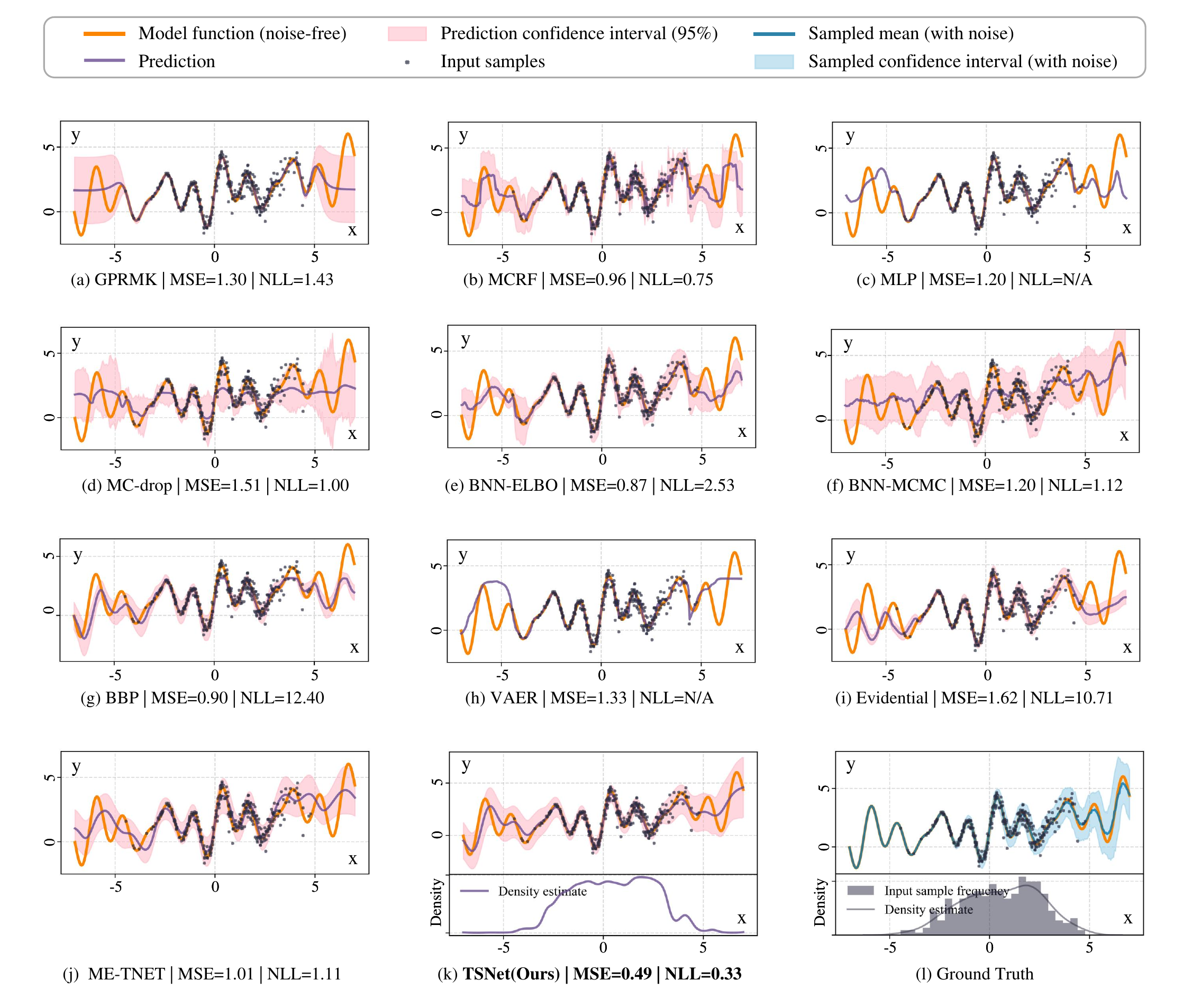}}
    \caption{Experimental results on the 1D toy example, covering interpolation and extrapolation predictions. (a)\textasciitilde(g) show model predictions, and (h) shows the true function. The models were trained on noisy data and evaluated on noise-free data using MSE. TSNet achieved the lowest MSE and accurately estimated uncertainty, reflecting the true noise and sampling characteristics.}
    \label{fig:exp1-1}
    \end{center}
\end{figure}

On the relatively large Year-Prediction-MSD dataset (514K samples), we assessed the time and GPU memory usage for both training and inference phases to confirm they were within acceptable limits. As shown in Table \ref{tab:exp_RU}, despite the low real-time requirements typically associated with scientific data tasks, the design of TSNet, particularly during the inference stage, allows for the retention of only the essential modules. This optimization enables model training times to be in the range of minutes and inference times to be approximately 1 second, with GPU memory usage consistently below 1GB, facilitating modeling on common GPUs.

We conducted scalability experiments on the Year-Prediction-MSD dataset by training models on training sets of varying sizes and observing changes in MSE. Additionally, we evaluated different model structures on the full training set of 463K samples. These experiments were performed three times with random shuffling and tested on the official test set of 51K samples, with results reported as averages and standard deviations.
The results, shown in Figures \ref{fig:exp5-6}, confirm that TSNet's accuracy improves with increasing data volume and model scale. This trend aligns with the typical performance of deep learning models, demonstrating TSNet's potential for handling large-scale data effectively. The experiments validate TSNet's scalability and efficiency, highlighting its suitability for large-scale scientific data applications.

\begin{figure}[t!]
    \begin{center}
    \centerline{\includegraphics[width=0.9\linewidth]{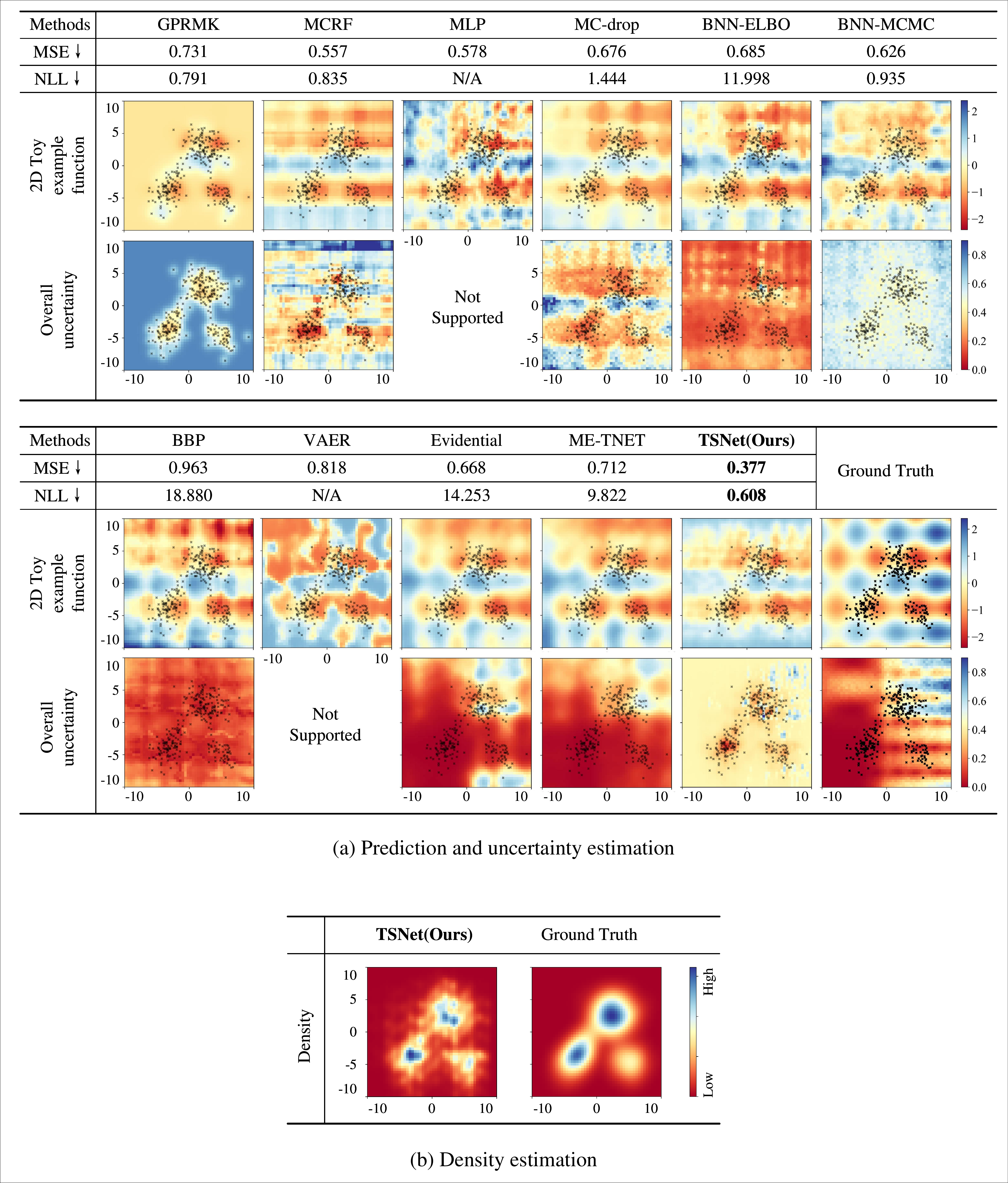}}
    \caption{Experimental results on the 2D toy function, covering interpolation and extrapolation predictions. Models were trained on noisy data and evaluated on noise-free data using MSE. TSNet achieved superior MSE results, estimating noise trends with limited data and providing comprehensive uncertainty estimates, taking into account noise patterns and sampling density.}
    \label{fig:exp1-2}
    \end{center}
\end{figure}

\subsection{Toy Example Experiments}
\label{sec:TEE}
Due to the challenge of obtaining noise-free scientific data, we designed toy experiments to reveal TSNet's process and demonstrate its noise-handling capabilities. These experiments serve to validate TSNet's ability to model uncertainty in both noisy and sparsely sampled environments, which are common in practical applications. The experiments included a 1D setup with one input and one output, and a 2D setup with two inputs and one output, providing a controlled environment to rigorously test the performance of TSNet.

\subsubsection{1D Toy Function Experiment} 

The function is expressed mathematically as
\begin{equation}
        \left\{
        \begin{array}{l}
        \begin{aligned}
        f_n(x_n) = & \exp(0.06x_{n}) + 0.5 \exp(0.2x_{n}) + \\
            & \sin(2x_{n}) + \sin(4x_{n}) + \sin(5x_{n}) + N_o \\
        x_{n} = & x + N_i 
        \end{aligned}
        \\
        N_i \sim \mathcal{N}(0, \sigma_i^2(x)), \\
        \sigma_i^2(x) = \left(0.2 \times \dfrac{1}{1+\exp(2-x)}\right)^2 \\
        N_o \sim \mathcal{N}(0, \sigma_o^2(x)), \\
        \sigma_o^2(x) = \left(0.5 \times \dfrac{1}{1-\exp(x+1)}\right)^2
        \end{array}.
        \right.
\end{equation}
When $N_i$=$N_o$=$0$, $f_n(x)$ reduces to the noise-free function $f(x)$. The visualization of $f_n(x)$ and $f(x)$ is shown in Figure \ref{fig:exp1-1} (l).
We sampled 300 noisy data points for training: 
\begin{equation}
    \begin{array}{l}
    \begin{aligned}
        \textbf{x}_{train1}^{(150)} \sim \mathcal{N}(-1,1.5^2\textbf{I}_{150})\\
        \textbf{x}_{train2}^{(150)} \sim \mathcal{N}(2.0,1.1^2\textbf{I}_{150})\\
    \end{aligned}
    \end{array},
\end{equation}
and generated 300 noise-free points uniformly distributed in the range $x \in [-7,7]$ for evaluation.

Figure \ref{fig:exp1-1} illustrates TSNet's superior MSE performance compared to other methods. MCRF and MLP showed tendencies to overfit in noisy regions, while GPRMK exhibited high uncertainty in extrapolated regions, leading to overly smooth predictions. MLP and VAER were overconfident and failed to capture data uncertainty. MC-dropout, BNN-ELBO, BNN-MCMC, Evidential, ME-TNET, and BBP could assess uncertainty and extrapolate but yielded suboptimal results.
However, TSNet effectively captured both aleatoric and epistemic uncertainties, accurately estimating noise distribution parameters and data sampling distributions. TSNet demonstrated a linear increase in predicted uncertainty with \(x\) in data-rich regions (e.g., \(x \in [-2,3]\)), reflecting noise variance. In data-sparse regions, TSNet extrapolated trends with increased uncertainty, showcasing its robustness in handling noisy and sparsely sampled data. From the evaluation metrics, TSNet outperformed other methods in both MSE and NLL.

This detailed analysis in 1D toy example experiments underscores its potential for real-world applications where data is often noisy and incomplete, highlighting its adaptability and reliability in uncertainty estimation.

\subsubsection{2D Toy Function Experiment} 
This experiment extends the complexity of our work by introducing a challenging scenario where the noise distribution parameters vary spatially and the noise variables are interdependent. The function used in this experiment is defined as
\begin{equation}
    \left\{
    \begin{array}{l}
    \begin{aligned}
    f_n(x_n,y_n) = & 0.5 \times \exp({0.03y_{n}}) \sin(x_{n}) + \\
            & \exp(0.06x_{n}) \cos(0.8y_{n}) + N_o \\
    x_{n} & = x + N_i \\
    y_{n} & = y + N_i 
    \end{aligned}
    \\
    N_i \sim \mathcal{N}(0, \Sigma_i(x,y)), \\
    \Sigma_i(x,y)=\left(\dfrac{0.2\sin(0.3y)+0.4}{1+\exp(-(x+1))}\right)^2 \\
    N_o \sim \mathcal{N}(0, \Sigma_o(x,y)), \\
    \Sigma_o(x,y)=\left(\dfrac{0.15\sin(0.3x)\sin(0.8y)+1.1}{1+\exp(2-y)}\right)^2   \\
    \end{array}.
    \right.
\end{equation}
$f(x,y)$ is the noise-free of $f_n(x,y)$, both visually represented in Figure \ref{fig:exp1-2} as the ground truth ``2D toy example function'' and ``overall uncertainty'', respectively. 
This 2D toy function creates a complex testing environment by incorporating spatially varying noise distribution parameters, thereby simulating real-world challenges.

The 2D toy experiment sampled 300 noisy data points from three different multivariate Gaussian distributions: 
\begin{equation}
    \begin{array}{l}
    \begin{aligned}
        \textbf{x}_{train1}^{(100 \times 2)} \! \sim \! \mathcal{N}(
        \begin{bmatrix}
          -2.9 \\
          -3.4
        \end{bmatrix}
        ,
        \begin{bmatrix}
          2.5 & 2.0 \\
          0.5 & 2.3
        \end{bmatrix}
        )\\

        \textbf{x}_{train2}^{(150 \times 2)} \! \sim \! \mathcal{N}(
        \begin{bmatrix}
          2.5 \\
          2.5
        \end{bmatrix}
        ,
        \begin{bmatrix}
          3.0 & 0.0 \\
          0.0 & 2.5
        \end{bmatrix}
        ) \\

        \textbf{x}_{train3}^{(50 \times 2)} \! \sim \! \mathcal{N}(
        \begin{bmatrix}
          5 \\
          -5
        \end{bmatrix}
        ,
        \begin{bmatrix}
          1.2 & -0.5 \\
          -0.7 & 1.7
        \end{bmatrix}
        )
    \end{aligned}
    \end{array}.
\end{equation}
Additionally, 2500 noise-free points were sampled in a $50 \! \times \! 50$ grid within $\{ (x,y) | -10 \! \le \! x \! \le \! 10, -10 \! \le \! y \! \le \! 10\}$ for evaluation purposes.

\begin{table}[t!]
    \caption{Results of the ablation experiment. The best results are shown in bold. ``In'', ``Out'', and ``All'' denote interpolation-MSE, extrapolation-MSE, and overall-MSE, respectively.}
    \label{tab:exp1}

    \begin{center}
    \resizebox{0.65\linewidth}{!}{
        \renewcommand{\arraystretch}{1.1}
        \begin{tabular}{ c | c | ccc | ccc }
        \hline

        \hline
        \multicolumn{2}{c|}{Toy example} & \multicolumn{3}{c|}{1D toy func (MSE$\downarrow$)} & \multicolumn{3}{c}{2D toy func (MSE$\downarrow$)} \\
        \hline
        \multicolumn{1}{@{}c@{}|}{Methods} & Module & In & Ext & All & In & Ext & All \\
        \hline
        \multirow{4}{*}{\thead{MLP\\based}} & MLP & 0.237 & 6.183 & 1.966 & 0.155 & 1.701 & 0.893 \\
        & + NCL & 0.405 & 2.224 & 0.925 & 0.136 & 1.258 & 0.688 \\
        & + DPN & 0.252 & 1.648 & 0.683 & 0.209 & 1.012 & 0.610 \\
        & + NCL + DPN & 0.226 & 1.546 & 0.674 & 0.131 & 0.942 & 0.568 \\
        \hline
        \multirow{4}{*}{TSNet} & DTB base & 0.163 & 2.091 & 0.657 & 0.175 & 1.025 & 0.533 \\
        & + NCL & 0.103 & 1.827 & 0.576 & 0.127 & 0.919 & 0.492 \\
        & + DPN & 0.149 & 1.280 & 0.558 & 0.180 & 0.672 & 0.388 \\
        & + NCL + DPN & \textbf{0.099} & \textbf{1.280} & \textbf{0.489} & \textbf{0.127} & \textbf{0.648} & \textbf{0.377} \\
        \hline

        \hline
        \end{tabular}
    }
    \end{center}

\end{table}

Figure \ref{fig:exp1-2} shows the superior performance of TSNet compared to other methods. 
Despite limited sampling, TSNet effectively anticipated increasing noise levels in the upper right quadrant $\{ (x,y) | -1 \! \le \! x \! \le \! 10, 2 \! \le \! y \! \le \! 10\}$ and maintained consistent uncertainty in undersampled regions.

These toy experiments highlight the difficulty of learning from data with unknown, spatially varying aleatoric uncertainty and limited, unevenly distributed points (i.e. epistemic uncertainty). TSNet's ability to estimate uncertainties consistent with noise and data patterns demonstrates its potential for experimental scientific applications.

\subsection{Ablation Experiments}
\label{sec:abexp}

Ablation studies were conducted to thoroughly evaluate the impact of TSNet's components on its predictive performance, with a specific focus on interpolation and extrapolation in scenarios with sparse and noisy data. These studies are crucial for understanding the contribution of each component to the overall performance of TSNet.
Following the model and test protocols described in section \ref{sec:TEE}, toy examples were utilized to assess the capabilities of interpolation, extrapolation, and overall data processing. The baseline model employed two separate MLPs for mean and uncertainty predictions.

Table \ref{tab:exp1} demonstrates that combining MLPs or DTB with NCL and DPM improves the overall MSE. Notably, the interplay between interpolation, extrapolation, and overall MSE is significant because it is challenging to simultaneously optimize all three metrics. The incorporation of NCL and DTB reduces overfitting in interpolation and enhances extrapolation, thereby validating the TSNet framework.

Moreover, replacing MLPs with DTB generally results in improved performance, underscoring the importance of DTB in capturing noise patterns. This improvement is attributed to the DTB's foundation in multivariate Taylor series theory and its design for heteroscedastic noise transformation, which enables it to effectively model complex noise distributions.

These ablation studies highlight the critical role of each component in TSNet, demonstrating that the integration of NCL and DPM with DTB significantly enhances the model's ability to handle noisy and sparse data, thereby validating the robustness and efficacy of the proposed TSNet framework.

\section{Limitations}

Although TSNet has achieved excellent results on real-world scientific datasets and challenging toy examples, further research is required to address its limitations.

\begin{itemize}
    \item The noise model defined by Equation (\ref{eq:noiseMod}) accounts for heteroscedastic multivariate probability distributions, which is sufficient for common scientific data scenarios. However, it is limited to zero-mean and additive noise. More complex scenarios that deviate from these assumptions may be captured by Deep Taylor Block's $F_{\Sigma_o}$, but the theoretical basis for this needs to be explored.
    \item The Deep Taylor Block uses first-order derivatives of $F_\mu$, which facilitates the use of automatic differentiation in modern deep learning frameworks. To ensure backward propagation of gradients, an additional requirement is proposed: $F_\mu$ must be twice differentiable.
\end{itemize}

\section{Conclusion}

In this paper, the Taylor-Sensus Network (TSNet) is proposed. It leverages a heteroscedastic noise transformation based on multivariate Taylor expansion to theoretically underpin its noise distribution estimation. TSNet incorporates DTB, NCL, DPM, and UCO, and explicitly considers data noise distribution and sampling density, providing interpretability to predictive uncertainty. Key contributions include the development of a novel heteroscedastic noise transformation and the introduction of a heteroscedasticity MSE loss, which enhance the model's ability to handle complex, noisy data. Multiple experiments have demonstrated the ability of TSNet to identify patterns and uncertainty in unevenly sampled and noisy data, outperforming state-of-the-art methods and providing valuable insights for data exploration.

\newpage

\appendix
\onecolumn

\section{Supplementary Derivations}

\subsection{Computation of Equation (\ref{eq:Taylor})}
\label{Apx:cmteqTaylor}
The multivariate Taylor series of a multivariable function $G(x)$ at $x_k$ is given by
\begin{equation}
\label{eq:taylor_proc}
    G(x) = G(x_k)+(\nabla G(x_k))^\top(x-x_k)+  \frac{1}{2!}(x-x_k)^\top \textbf{H}_G(x_k)(x-x_k)+o^n,
\end{equation}
For the multivariate Taylor series of $F(\textbf{X} + N_i(\textbf{X}))$ at $\textbf{X}$, set
\begin{equation}
    x = \textbf{X} + N_i(\textbf{X}), \quad x_k = N_i(\textbf{X}),
\end{equation}
and substituting into Equation (\ref{eq:taylor_proc}) yields Equation (\ref{eq:Taylor}).

\subsection{Derivations of Equation (\ref{eq:gasdis})}
\label{apx:cmptEqMultigas}

Given that
\begin{equation}
    N_i(\textbf{X}) \sim \mathcal{N}(0, \Sigma_i(\textbf{X})),
\end{equation}
\begin{equation}
    N_o(\textbf{X}) \sim \mathcal{N}(0, \Sigma_o(\textbf{X})).
\end{equation}
The moment-generating functions (MGFs) for $N_i(\textbf{X})$ and $N_o(\textbf{X})$ are respectively:
\begin{equation}
\label{eq:tmpmutigastmp11}
    M_{N_i}(t) = \exp\left( \frac{1}{2} t^\top \Sigma_i(\textbf{X}) t \right),
\end{equation}
\begin{equation}
\label{eq:tmpmutigastmp22}
    M_{N_o}(t) = \exp\left( \frac{1}{2} t^\top \Sigma_o(\textbf{X}) t \right).
\end{equation}
Considering the linear combination
\begin{equation}
    N(\textbf{X})=( \nabla F(\textbf{X}) ) ^\top N_i(\textbf{X}) + N_o(\textbf{X}),
\end{equation}
we compute the MGF of $N(\textbf{X})$:
\begin{equation}
\label{eq:tmpmutigastmp}
\begin{aligned}
    M_{N}(t) =& \mathbb{E} \left[ \exp(t^\top N(\textbf{X})) \right] \\
            =& \mathbb{E} \left[ \exp(t^\top ((\nabla F(\textbf{X}))^\top N_i(\textbf{X}) + N_o(\textbf{X}))) \right] \\
            =& \mathbb{E} \left[ \exp(t^\top (\nabla F(\textbf{X}))^\top N_i(\textbf{X})) \cdot \exp(t^\top N_o(\textbf{X})) \right] \\
            =& \mathbb{E} \left[ \exp(t^\top (\nabla F(\textbf{X}))^\top N_i(\textbf{X})) \right] \cdot \mathbb{E} \left[ \exp(t^\top N_o(\textbf{X})) \right]
\end{aligned}.
\end{equation}
According to Equations (\ref{eq:tmpmutigastmp11}) and (\ref{eq:tmpmutigastmp22}) for the MGFs of $N_i(\textbf{X})$ and $N_o(\textbf{X})$, we can express the expectations in Equation (\ref{eq:tmpmutigastmp}) using the MGFs of $N_i(\textbf{X})$ and $N_o(\textbf{X})$,
\begin{equation}
\begin{aligned}
M_{N}(t) = & M_{N_i}((\nabla F(\textbf{X}))^\top t) \cdot M_{N_o}(t) \\
        = & \exp\left( \frac{1}{2} ((\nabla F(\textbf{X}))^\top t)^\top \Sigma_i(\textbf{X}) ((\nabla F(\textbf{X}))^\top t) \right) \cdot \exp\left( \frac{1}{2} t^\top \Sigma_o(\textbf{X}) t \right) \\
        = & \exp\left( \frac{1}{2} t^\top ((\nabla F(\textbf{X}))^\top \Sigma_i(\textbf{X}) (\nabla F(\textbf{X})) + \Sigma_o(\textbf{X})) t \right)
\end{aligned}.
\end{equation}
Based on the equivalence of MGFs and probability density functions, we can conclude that $N(\textbf{X})$ follows a multivariate Gaussian distribution with a mean of zero and a covariance of $(\nabla F(\textbf{X}))^\top \Sigma_i(\textbf{X}) (\nabla F(\textbf{X})) + \Sigma_o(\textbf{X})$.

\subsection{Derivations of Equation (\ref{eq:finallres})}
\label{apx:cmptfinallres}
Assuming the random variables $N_n(\textbf{X})$ and $N_p(\textbf{X})$ with heteroscedasticity are independent and follow a multivariate Gaussian distribution
\begin{equation}
    N_n(\textbf{X}) \sim \mathcal{N}(\mu_n(\textbf{X}), \Sigma_n(\textbf{X})),
\end{equation}
\begin{equation}
    N_p(\textbf{X}) \sim \mathcal{N}(\mu_p(\textbf{X}), \Sigma_p(\textbf{X})).
\end{equation}
Given density-sensitive weight $k_d(\textbf{X})$, consider a new random variable $N_c(\textbf{X})=k_d(\textbf{X})N_n(\textbf{X})+(1-k_d(\textbf{X}))N_p(\textbf{X})$. Calculate the distribution parameters of $N_c(\textbf{X})$.

The moment-generating functions (MGFs) for $N_n(\textbf{X})$ and $N_p(\textbf{X})$ are respectively:
\begin{equation}
\label{eq:mnn}
    M_{N_n}(t) = \exp \left( t^\mathrm{T} \mu_n(\textbf{X}) + \frac{1}{2} t^\mathrm{T} \Sigma_n(\textbf{X}) t \right),
\end{equation}
\begin{equation}
\label{eq:mnp}
    M_{N_p}(t) = \exp \left( t^\mathrm{T} \mu_p(\textbf{X}) + \frac{1}{2} t^\mathrm{T} \Sigma_p(\textbf{X}) t \right).
\end{equation}
Compute the MGF of $N_c(\textbf{X}) = k_d(\textbf{X})N_n(\textbf{X}) + (1 - k_d(\textbf{X}))N_p(\textbf{X})$:
\begin{equation}
\begin{aligned}
    M_{N}(t) = & \mathbb{E} \left[ \exp \left( t^\mathrm{T} \left( k_d(\textbf{X})N_n(\textbf{X}) + (1 - k_d(\textbf{X}))N_p(\textbf{X}) \right) \right) \right] \\
    = & \mathbb{E} \left[ \exp \left( t^\mathrm{T} k_d(\textbf{X})N_n(\textbf{X}) \right) \cdot \exp \left( t^\mathrm{T} (1 - k_d(\textbf{X}))N_p(\textbf{X}) \right) \right] \\
    = & \mathbb{E} \left[ \exp \left( t^\mathrm{T} k_d(\textbf{X})N_n(\textbf{X}) \right) \right] \cdot \mathbb{E} \left[ \exp \left( t^\mathrm{T} (1 - k_d(\textbf{X}))N_p(\textbf{X}) \right) \right]
\end{aligned}.
\end{equation}
Use the MGFs of $N_n(\textbf{X})$ and $N_p(\textbf{X})$ to represent these expectations:
\begin{equation}
    M_N(t) = M_{N_n}(k_d(\textbf{X})t) \cdot M_{N_p}((1 - k_d(\textbf{X}))t).
\end{equation}
Substitute the Equations (\ref{eq:mnn}) and (\ref{eq:mnp}):
\begin{equation}
\begin{aligned}
\label{eq:mnt}
    M_N(t) = & \exp \left( (k_d(\textbf{X})t)^\mathrm{T} \mu_n + \frac{1}{2} (k_d(\textbf{X})t)^\mathrm{T} \Sigma_n (k_d(\textbf{X})t) \right) \cdot \\
     & \exp \left( ((1 - k_d(\textbf{X}))t)^\mathrm{T} \mu_p + \frac{1}{2} ((1 - k_d(\textbf{X}))t)^\mathrm{T} \Sigma_p ((1 - k_d(\textbf{X}))t) \right) \\
    = & \exp \left( k_d(\textbf{X}) t^\mathrm{T} \mu_n + \frac{1}{2} k_d(\textbf{X})^2 t^\mathrm{T} \Sigma_n t \right) \cdot \exp \left( (1 - k_d(\textbf{X})) t^\mathrm{T} \mu_p + \frac{1}{2} (1 - k_d(\textbf{X}))^2 t^\mathrm{T} \Sigma_p t \right)
\end{aligned}.
\end{equation}
By comparing Equations (\ref{eq:mnt}) with the MGF of a standard multivariate Gaussian distribution:
\begin{equation}
    M_y(t) = \exp \left( t^\mathrm{T} \mu + \frac{1}{2} t^\mathrm{T} \Sigma t \right).
\end{equation}
Since $N_c(x) \sim \mathcal{N} (\mu, \Sigma)$,
\begin{equation}
\mu = k_d \mu_n + (1 - k_d) \mu_p,
\end{equation}
\begin{equation}
\Sigma = k_d^2 \Sigma_n + (1 - k_d)^2 \Sigma_p + 2k_d(1 - k_d) \text{Cov}(N_n, N_p).
\end{equation}
Since $N_n(x)$ and $N_p(x)$ are independent, $\text{Cov}(N_n, N_p)$ is the zero matrix,
\begin{equation}
\Sigma = k_d^2 \Sigma_n + (1 - k_d)^2 \Sigma_p.
\end{equation}

\subsection{Derivations of Equation (\ref{eq:HMSE})}
\label{apx:cmptHMSE}
The probability density function (PDF) of the multivariate Gaussian distribution is
\begin{equation}
    f(y|\widetilde{y}, \Sigma) = \frac{1}{\sqrt{(2\pi)^k |\Sigma|}} \exp\left(-\frac{1}{2} (y-\widetilde{y})^\top \Sigma^{-1} (y-\widetilde{y})\right).
\end{equation}
The log-likelihood function is
\begin{equation}
\log p(y| \widetilde{y}, \Sigma) = -\frac {1}{2} \log ((2 \pi)^k |\Sigma|)-\frac{1}{2}(y-\widetilde {y})^ \top \Sigma ^{-1} (y-\widetilde {y}).
\end{equation}
For constructing the loss function, we consider the negative log-likelihood, leading to the following expression:
\begin{equation}
\begin{aligned}
    \mathcal{L}_{HMSE} = & \frac{1}{2} \sum^{k}_{i=1} (\log((2 \pi)^k |\Sigma|) + (y - \widetilde{y})^T \Sigma^{-1} (y - \widetilde{y})) \\
    \simeq & \sum^{k}_{i=1} (\log(|\Sigma|) + (y - \widetilde{y})^T \Sigma^{-1} (y - \widetilde{y}))
\end{aligned}.
\end{equation}

\section{The Validity of Model Assumptions}

To ensure the robustness and applicability of TSNet in real-world scenarios, we explore the scientificity of the underlying model assumptions through both theoretical analysis and empirical experiments.

\subsection{Heteroscedastic Multivariate Gaussian's Expressive Power}

The heteroscedastic multivariate Gaussian distribution \(\mathcal{N}(\mu(X), \Sigma(X))\) is employed to accommodate local data characteristics, offering adaptable Gaussian approximations. The Central Limit Theorem indicates that the mean of a sufficiently large number of independent, identically distributed random variables with finite mean and variance converges to a normal distribution. Consequently, the conditional distributions of certain non-Gaussian processes can be effectively modeled by Gaussian distributions, such as the Gamma distribution. Additionally, real-world noise often arises from the accumulation of multiple sources of random noise that satisfy the Central Limit Theorem and exhibit Gaussian characteristics, thus supporting our basic noise assumption.

\subsection{Generalized Noise Decomposition Model}
The Deep Taylor Block (DTB, Equation (\ref{eq:DTB})) is designed to address the specified noise model $Y_n=F(X+N_i)+N_o$ (Equation (\ref{eq:noiseMod})). In the generalization of a more comprehensive noise model:
\begin{equation}
    Y_n=Noise(F(X)),
\end{equation}
where $Noise(\cdot)$ represents any noise influencing the ideal noiseless system $F(X)$. Under this formulation, the noise, represented by the discrepancy between observed and ideal values $Noise(F(X))-F(X)$, remains associated with the system $F$.

Under optimal conditions, a sufficiently expressive $F_{\Sigma_o}$ network can accommodate any noise term $Noise(F(X))-F(X)$. In real-world applications, however, accurately learning noise from limited data poses a significant challenge. Enhancing noise estimation requires the inclusion of suitable priors for the noise type during training, such as designing the DTB structure in accordance with the noise model and the assumption of zero-mean Gaussian noise.

\subsection{Deviation from Noise Assumptions}

\begin{figure}[t!]
    \begin{center}
    \centerline{\includegraphics[width=\linewidth]{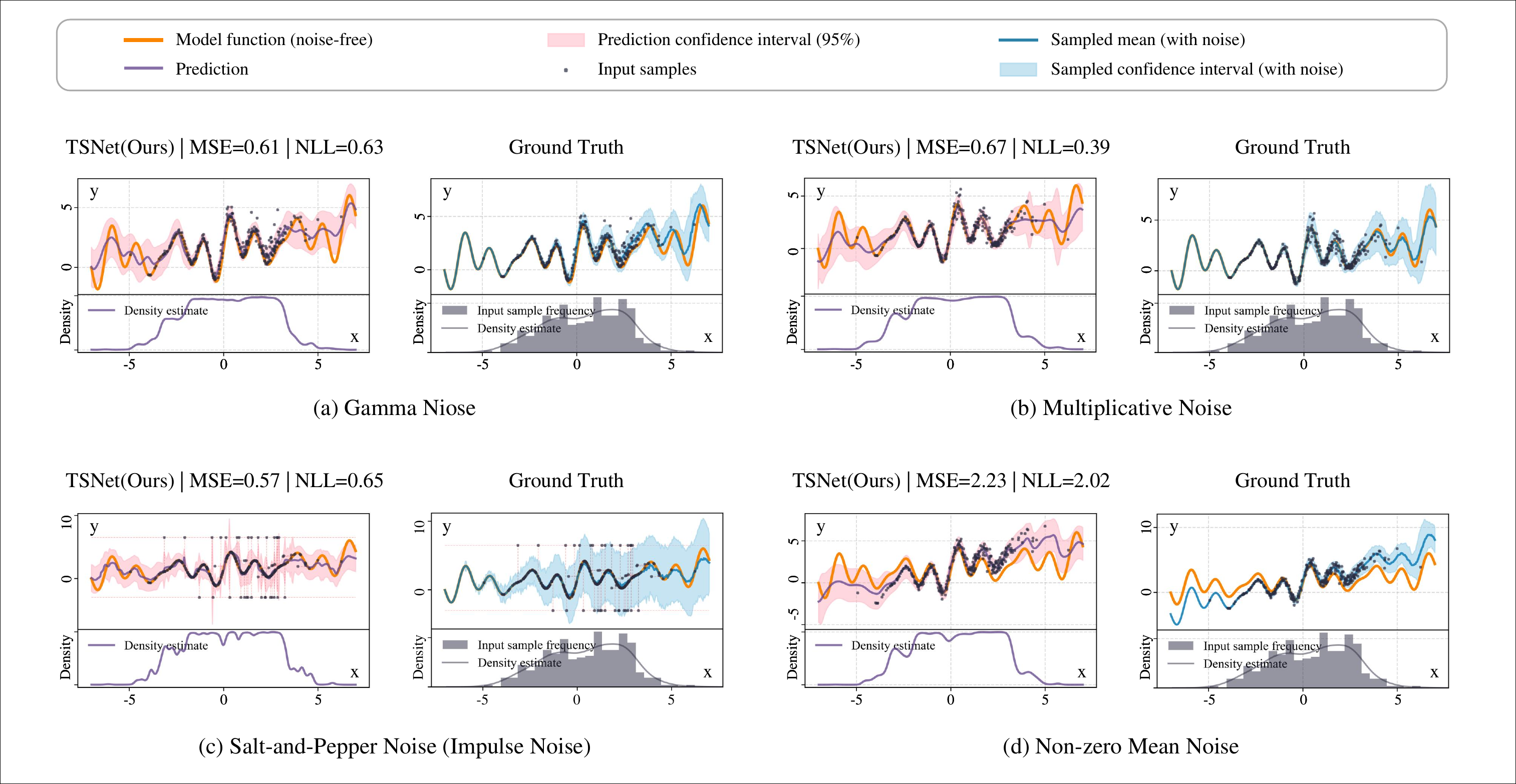}}
    \caption{Experiments with Diverse Noise. (a)\textasciitilde(d) show the results of introducing various types of noise.}
    \label{fig:exp5-9}
    \end{center}
\end{figure}

To evaluate the robustness of TSNet to deviations from the Gaussian noise assumption, we conducted a series of experiments introducing diverse noise types.
In Figure \ref{fig:exp5-9}(a), Gamma noise exhibits formal similarities to Gaussian noise, with heteroscedastic Gaussian noise proving capable of adequate prediction. 
Figure \ref{fig:exp5-9}(b) demonstrates that DTB maintains robust noise handling capabilities in the presence of multiplicative noise. 
Figure \ref{fig:exp5-9}(c) shows salt-and-pepper noise, which diverges significantly from the Gaussian pattern, often resembling missing data, outliers, and high nonlinearity, highlighting TSNet's ability to capture such noise characteristics. 
Figure \ref{fig:exp5-9}(d) reveals a limitation of TSNet in capturing the mean shift of additive non-zero mean Gaussian noise. Despite the representability of additive noise by \(F_{\Sigma_o}\), the model's inability to discern noise from actual data due to insufficient prior information precludes successful learning of the noise pattern.

Although \(F_{\Sigma_o}\) has the potential to capture noise patterns for more complex or generalized noise conditions, specific conditions and theoretical derivations require further exploration. Overall, these experiments validate the robustness of TSNet under the noise scenarios that deviate from the noise assumptions, while identifying areas for future improvement.

\begin{table}[t!]
\caption{Computational efficiency experiments results for scientific datasets. ``Concrete Compressive Strength'' is abbreviated as ``Concrete Comp. Str.''. The results are reported as 'mean ± standard deviation'.}
\label{tab:exp2r}
\begin{center}
\resizebox{0.85\linewidth}{!}{

\renewcommand{\arraystretch}{1.2}

\begin{tabular}{ c | r@{±}l r@{±}l | r@{±}l r@{±}l | r@{±}l r@{±}l }
\hline

\hline
\multirow{2}{*}{Datasets} & \multicolumn{4}{c|}{Matbench steels} & \multicolumn{4}{c|}{Diabetes} & \multicolumn{4}{c}{Forest fires} \\
\cline{2-13}
 & \multicolumn{2}{c}{Training set} & \multicolumn{2}{c|}{Testing set} & \multicolumn{2}{c}{Training set} & \multicolumn{2}{c|}{Testing set} & \multicolumn{2}{c}{Training set} & \multicolumn{2}{c}{Testing set} \\
 \hline
Sample count & \multicolumn{2}{c}{187} & \multicolumn{2}{c|}{63} & \multicolumn{2}{c}{265} & \multicolumn{2}{c|}{89} & \multicolumn{2}{c}{310} & \multicolumn{2}{c}{104} \\
\hline
Methods & \multicolumn{4}{c|}{Total time (s)} & \multicolumn{4}{c|}{Total time (s)} & \multicolumn{4}{c}{Total time (s)} \\
\hline
GPRMK & 0.016 & 0.004 & 0.008 & 0.004 & 0.020 & 0.008 & 0.037 & 0.012 & 0.025 & 0.006 & 0.014 & 0.003 \\
MCRF & 0.257 & 0.014 & 0.170 & 0.014 & 0.244 & 0.004 & 0.192 & 0.001 & 0.246 & 0.001 & 0.198 & 0.001 \\
MLP & 3.177 & 0.000 & 0.001 & 0.000 & 2.394 & 0.015 & 0.002 & 0.000 & 2.987 & 0.041 & 0.002 & 0.000 \\
MC-drop & 2.723 & 0.009 & 0.863 & 0.009 & 2.762 & 0.033 & 1.061 & 0.001 & 3.409 & 0.196 & 1.065 & 0.003 \\
BNN-ELBO & 9.591 & 0.235 & 3.818 & 0.235 & 8.883 & 0.328 & 4.031 & 0.233 & 10.838 & 0.110 & 4.204 & 0.010 \\
BNN-MCMC & 165.172 & 0.122 & 1.255 & 0.122 & 143.668 & 8.299 & 1.072 & 0.017 & 264.917 & 53.682 & 1.479 & 0.277 \\
BBP & 72.963 & 0.025 & 4.090 & 0.025 & 68.333 & 3.151 & 4.636 & 0.018 & 84.860 & 1.217 & 4.827 & 0.088 \\
VAER & 6.202 & 0.000 & 0.002 & 0.000 & 5.061 & 0.545 & 0.002 & 0.000 & 5.165 & 0.322 & 0.002 & 0.000 \\
Evidential & 4.647 & 0.000 & 0.002 & 0.000 & 3.255 & 0.015 & 0.002 & 0.000 & 4.461 & 0.120 & 0.002 & 0.000 \\
MT-ENet & 5.779 & 0.000 & 0.002 & 0.000 & 4.164 & 0.049 & 0.002 & 0.000 & 5.914 & 0.558 & 0.003 & 0.000 \\
\hline
\textbf{TSNet(Ours)} & 14.516 & 0.001 & 0.006 & 0.001 & 14.176 & 0.011 & 0.007 & 0.000 & 13.522 & 0.484 & 0.005 & 0.000\\
\hline

\hline
\end{tabular}
}

\vspace{0.2cm}

\resizebox{0.85\linewidth}{!}{

\renewcommand{\arraystretch}{1.2}

\begin{tabular}{ c | r@{±}l r@{±}l | r@{±}l r@{±}l | r@{±}l r@{±}l }
\hline

\hline
\multirow{2}{*}{Datasets} & \multicolumn{4}{c|}{Concrete
Comp. Str.} & \multicolumn{4}{c|}{Wine quality-red} & \multicolumn{4}{c}{Wine quality-white} \\
\cline{2-13}
 & \multicolumn{2}{c}{Training set} & \multicolumn{2}{c|}{Testing set} & \multicolumn{2}{c}{Training set} & \multicolumn{2}{c|}{Testing set} & \multicolumn{2}{c}{Training set} & \multicolumn{2}{c}{Testing set} \\
 \hline
Sample count & \multicolumn{2}{c}{618} & \multicolumn{2}{c|}{206} & \multicolumn{2}{c}{959} & \multicolumn{2}{c|}{320} & \multicolumn{2}{c}{2939} & \multicolumn{2}{c}{979} \\
\hline
Methods & \multicolumn{4}{c|}{Total time (s)} & \multicolumn{4}{c|}{Total time (s)} & \multicolumn{4}{c}{Total time (s)} \\
\hline
GPRMK & 0.060 & 0.008 & 0.192 & 0.008 & 0.170 & 0.031 & 0.198 & 0.013 & 0.995 & 0.045 & 0.366 & 0.005 \\
MCRF & 0.361 & 0.002 & 0.188 & 0.002 & 0.723 & 0.016 & 0.199 & 0.002 & 1.994 & 0.048 & 0.227 & 0.002 \\
MLP & 4.773 & 0.000 & 0.002 & 0.000 & 8.078 & 0.084 & 0.003 & 0.000 & 28.690 & 0.114 & 0.005 & 0.000 \\
MC-drop & 5.596 & 0.022 & 1.318 & 0.022 & 8.343 & 0.368 & 1.869 & 0.098 & 32.324 & 1.509 & 4.335 & 0.112 \\
BNN-ELBO & 11.778 & 0.012 & 3.984 & 0.012 & 12.596 & 0.507 & 3.902 & 0.276 & 15.419 & 1.550 & 4.054 & 0.235 \\
BNN-MCMC & 275.096 & 0.187 & 1.471 & 0.187 & 272.238 & 21.963 & 1.686 & 0.004 & 644.075 & 98.228 & 3.041 & 1.023 \\
BBP & 90.009 & 0.032 & 4.017 & 0.032 & 108.872 & 6.493 & 4.767 & 0.070 & 303.083 & 0.371 & 4.710 & 0.018 \\
VAER & 9.845 & 0.000 & 0.003 & 0.000 & 12.439 & 0.652 & 0.003 & 0.000 & 37.944 & 1.735 & 0.009 & 0.001 \\
Evidential & 6.085 & 0.000 & 0.002 & 0.000 & 9.550 & 0.248 & 0.003 & 0.000 & 35.774 & 2.975 & 0.005 & 0.000 \\
MT-ENet & 8.239 & 0.000 & 0.003 & 0.000 & 11.562 & 0.169 & 0.004 & 0.001 & 37.262 & 0.766 & 0.006 & 0.001 \\
\hline
\textbf{TSNet(Ours)} & 21.466 & 0.000 & 0.007 & 0.000 & 25.208 & 0.097 & 0.007 & 0.000 & 77.700 & 0.425 & 0.009 & 0.000 \\
\hline

\hline
\end{tabular}

}
\end{center}
\end{table}

\section{Experiment Details}

\subsection{Experiment Setting Details}
For each dataset in experiments, all deep learning methods (MLP, MC-drop, BNN-ELBO, BNN-MCMC, BBP, VAER, Evidental, MT-ENet, and TSNet) employed a consistent main network architecture. Additionally, a grid search for hyperparameter optimization was conducted for all models, ensuring a fair comparison.

In the toy example experiments (Section \ref{sec:TEE}) and the ablation experiments (Section \ref{sec:abexp}), 1D and 2D toy examples were designed for easier visualization. To enhance input features and ensure fairness, positional embedding was applied to all features in all of the toy example experiments:
\begin{equation}
    \textbf{X}_\text{emd} = (\textbf{X}, \sin(2^0 \pi \textbf{X}), \cos(2^0 \pi \textbf{X}), \sin(2^1 \pi \textbf{X}), \cos(2^1 \pi \textbf{X}), \ldots, \sin(2^L \pi \textbf{X}), \cos(2^L \pi \textbf{X})),
\end{equation}
where $L$ is a hyperparameter.

\subsection{Scientific Datasets Preprocessing}
\label{apx:data_pre}

We followed standard ML practice by applying the same preprocessing to the data across all experiments, ensuring fairness among the methods benchmarked. Since this work is not domain-specific, no further feature engineering was conducted.
\begin{itemize}
    \item \textbf{Matbench Steels} \citep{dunn2020benchmarking}. This dataset from materials science was standardized for input features and labels before model input.
    \item \textbf{Diabetes} \citep{efron2004least}. The medical dataset was standardized for input features and labels before modeling.
    \item \textbf{Forest Fires} \citep{misc_forest_fires_162}. For this environmental science dataset, ``month'' and ``day'' were mapped to numbers 1\textasciitilde12 and 1\textasciitilde7, respectively, and features ``rain'' and ``area'' with long-tailed distributions were log-transformed. Input features and labels were then standardized.
    \item \textbf{ Concrete Compressive Strength} \citep{misc_concrete_compressive_strength_165}. This materials science dataset was standardized for input features and labels.
    \item \textbf{Wine (red/white) Quality} \citep{misc_wine_quality_186}. 
    Originally a classification task for ``quality'' in food science, it was adapted to a regression task for ``density'' to align with our work’s objectives. These datasets inherently involve a degree of subjectivity in quality assessment.
    \begin{itemize}
        \item \textbf{Wine Quality-red}. features with long-tailed distributions, including ``residual sugar'', ``chlorides'', ``free sulfur dioxide'', ``total sulfur dioxide'', and ``sulphates'', were log-transformed. Input features and labels were standardized.
        \item \textbf{Wine Quality-white}. features with long-tailed distributions, including ``fixed acidity'', ``volatile acidity'', ``residual sugar'', ``chlorides'', ``free sulfur dioxide'', ``total sulfur dioxide'', and ``density'', were log-transformed. Input features and labels were standardized.
    \end{itemize}    
\end{itemize}

\subsection{Computational Efficiency Experiments}

To evaluate the computational efficiency of TSNet, we performed training and inference time statistics on a single NVIDIA 1080Ti GPU across scientific datasets. 
As shown in Table \ref{tab:exp2r}, despite TSNet being tailored for scientific data processing, where high real-time performance is typically not a primary requirement, it demonstrates competitive efficiency in terms of computational resources, particularly during the inference phase.

During the training phase, TSNet outperformed both BNN-MCMC and BBP, requiring a maximum of 77 seconds to train on the Wine quality-white dataset. This showcases TSNet's relatively rapid training capabilities. In the inference phase, TSNet's streamlined architecture allowed it to complete all test set predictions within milliseconds, underscoring its efficiency in real-time applications.
These results show that TSNet's computational efficiency is acceptable and suitable for scientific research applications, balancing robust performance with resource-effective processing.

\newpage

\bibliographystyle{plainnat}
\bibliography{references}

\end{document}